\documentclass[11pt,3p,times,authoryear]{elsarticle}

\usepackage{amssymb}
\usepackage{booktabs}
\usepackage{amsmath}
\usepackage{lipsum}
\usepackage{subcaption}
\usepackage{algpseudocode}
\usepackage{algorithm}
\usepackage{hyperref}
\usepackage[capitalise]{cleveref}
\crefformat{appendix}{#2#1#3}
\usepackage{tikz}
\usetikzlibrary{positioning,arrows.meta,fit,backgrounds}

\definecolor{InputBlue}{RGB}{52, 152, 219}
\definecolor{AlgoOrange}{RGB}{230, 126, 34}
\definecolor{EstGreen}{RGB}{39, 174, 96}
\definecolor{LatentPurple}{RGB}{155, 89, 182}

\usepackage[nolist]{acronym}
\usepackage{url}

\journal{RESS}

\begin{document}

\begin{frontmatter}





\title{A Bayesian Learning Approach for Drone Coverage Network: A Case Study on Cardiac Arrest in Scotland}


\author[1]{T. Basu}
\author[2]{E. Patelli\corref{cor1}}
\author[3]{G. Filippi}
\author[4]{B. Parsonage}
\author[4]{C. Maddock}
\author[4]{M. Vasile}
\author[4]{M. Fossati}
\author[5]{A. Loyd}
\author[5]{S. Marshall}
\author[5]{P. Gowens}

\affiliation[1]{{UNSW Bengaluru, Bengaluru, India}}
\affiliation[2]{{Civil and Environmental Engineering, University of Strathclyde, Glasgow G1 1XQ, UK}}
\affiliation[3]{{Centre for Maritime Research and Experimentation, 19126 La Spezia, Italy}}
\affiliation[4]{{Mechanical and Aerospace Engineering, University of Strathclyde, Glasgow G1 1XQ, UK}}
\affiliation[5]{{Scottish Ambulance Service, UK}}
\cortext[cor1]{edoardo.patelli@strath.ac.uk}

\begin{abstract}
	Drones are becoming popular as a complementary system for \ac{ems}. Although several pilot studies and flight trials have shown the feasibility of drone-assisted \ac{aed} delivery, running a full-scale operational network remains challenging due to high capital expenditure and environmental uncertainties. In this paper, we formulate a reliability-informed Bayesian learning framework for designing drone-assisted \ac{aed} delivery networks under environmental and operational uncertainty. We propose our objective function based on the survival probability of \ac{ohca} patients to identify the ideal locations of drone stations. Moreover, we consider the coverage of existing \ac{ems} infrastructure to improve the response reliability in remote areas. We illustrate our proposed method using geographically referenced cardiac arrest data from Scotland. The result shows how environmental variability and spatial demand patterns influence optimal drone station placement across urban and rural regions. In addition, we assess the robustness of the network and evaluate its economic viability using a cost-effectiveness analysis based on expected \ac{qaly}. The findings suggest that drone-assisted \ac{aed} delivery is expected to be cost-effective and has the potential to significantly improve the emergency response coverage in rural and urban areas with longer ambulance response times.
	
\end{abstract}


\end{frontmatter}


\begin{acronym}[MPC]
\acro{aed}[AED]{Automated External Defibrillator}

\acro{ems}[EMS]{Emergency Medical Service}

\acro{ga}[GA]{Genetic Algorithm}
\acro{ggc}[GGC]{Greater Glasgow and Clyde}
\acro{gis}[GIS]{Geographic Information System}
\acro{gp}[GP]{Gaussian Process}

\acro{kpi}[KPI]{Key Performance Indicator}

\acro{mclp}[MCLP]{Maximum Coverage Location Problem}

\acro{nfz}[NFZ]{No Fly Zone}
\acro{nhs}[NHS]{National Health Service}
\acro{nice}[NICE]{National Institute for Health and Care Excellence}

\acro{ohca}[OHCA]{Out of Hospital Cardiac Arrest}

\acro{pad}[PAD]{Public Access Defibrillator}

\acro{qaly}[QALY]{Quality Adjusted Life Year}

\acro{sas}[SAS]{Scottish Ambulance Service}

\acro{uav}[UAV]{Unmanned Aerial Vehicle}
\acro{ucg}[UCG]{Utstein Comparator Group}

\acro{vtol}[VTOL]{Vertical Take-off and Landing}
\end{acronym}

\section{Introduction}

Coronary artery disease remains one of the leading causes of death in Scotland according to Public Health Information for Scotland. In 2023, 12.4\% of all deaths (22,857) under the age of 75 years were directly related to coronary artery disease, and each year more than 3000 people experience \ac{ohca} \citep{sas_ohca_report}. Epidemiological studies consistently demonstrate that timely intervention is critical for survival, highlighting the importance of rapid and reliable emergency response systems \citep{larsen1993,valenzuela1997}. Although existing \ac{ems} has a structured response network, conventional ground-based ambulance deployment is constrained by road congestion, geographical remoteness, and crew availability, leading to severe uncertainty in response time and reduced reliability in timely intervention.

Recent advances suggest that \ac{uav} technology can complement existing \ac{ems} infrastructure by enabling rapid \ac{aed} delivery. Empirical studies and trial-based investigations demonstrate that drone-assisted \ac{aed} delivery can reduce response time and potentially improve survival outcomes \citep{Pulver2016,boutilier_2017_optimizing,RYAN2021532,Schierbeck2021,LEUNG202224,BAUMGARTEN2022139,schierbeck_2023_drone,STARKS2024101033,Smith2025}. Economic assessments further indicate that drone deployments can be cost-effective under suitable network planning \citep{Bauere043791,Roper2023,Maaz2025}. However, large-scale implementation remains challenged by operational, regulatory, and public integration constraints as discussed by \citet{ZegreHemsey2024}. {Furthermore, the treatment of epistemic and aleatory uncertainty in the network design, due to different external factors remains unaddressed which makes it a difficult decision to make for the policy makers.}

In this paper, we aim to address this challenge by proposing a reliability informed framework for understanding the viability of \ac{uav} assisted \ac{aed} deliveries in Scotland. In general, the network design {for \ac{uav} assisted \ac{aed} delivery} can be formulated as a location--allocation problem \citep{Cooper1963}, where we seek to find optimal locations for facilities to cover different demand areas. In the earlier works on the topic, most studies were conducted with fixed distance/travel time, binary coverage definitions, and deterministic demand patterns  \citep{CHURCH1974101,Narula1977,Rosing1979,Kariv1979}. Later on, stochastic extensions were proposed for scenario-based uncertainty and disruption modelling \citep{Snyder01062006,BERMAN200819,AN201454} to robustify the design. Similarly, from the engineering perspective, several works have been done to investigate the structural and operational reliability of \ac{uav} systems at the component and network level \citep{XU2022108221,Xing2023,Zaitseva2023,Ren2025}, highlighting the need for a network level performance evaluation. On the other hand, \citet{Stolaroff2018,Sorbelli2023,Silva2024} studied the flight performance under different environmental scenarios for a better design of drone carriages and controller system.

{Despite these progresses in the field, three major practical issues persist when covering problem is considered for designing a reliable drone delivery network. First of all, drone flight time is a stochastic entity that heavily depends on the the favourable wind conditions that has a spatio-temporal effect; scenario-based models can capture some of this variability but do not propagate it into the coverage probability in a principled manner. Secondly, standard formulations only rely on the travel time for mission feasibility and  do not account for the energy consumption, a constraint specific to battery run-drones. Lastly, demand factor in existing models is typically fixed or, scenario-based, and does not count for the uncertainty in ambulance response times.}

{
\subsection{Our Contribution}

We address these limitations by formulating the entire network-design problem as a Bayesian decision problem. Instead of solving an optimisation problem, we obtain the network design by posterior inference over the space of admissible designs. This is the main conceptual foundation of our contribution. Our approach yields a full posterior over designs, from which design uncertainty, robustness, and tail behaviour follow directly without needing to perform any post-hoc assessment.

In recent years, Bayesian methods have gained significant attention for uncertainty quantification in engineering systems. Gaussian Process (GP) regression and Bayesian calibration are now standard tools for uncertainty propagation and the calibration of complex simulators \citep{NIPS1995_GP,rasmussen2005gaussian,Kennedy2001,OHagan2006}, and Bayesian reliability analysis has become increasingly prominent \citep{Pan2009,Walter2017,Wilson2017,Leoni2021}, owing to its ability to combine prior structural knowledge with observed data and to quantify uncertainty under limited information. Our approach differs from conventional Bayesian methods that impose distributional assumptions on a likelihood function, and advances the available methods in three principal and clearly delineated aspects.

Firstly, we model coverage probabilistically through response-time distributions rather than deterministic distance or travel-time functions, which allows environmental variability and system reliability to enter the definition of service effectiveness in a single coherent construction. Specifically, we formulate a loss function corresponding to a parallel reliability system, a form that applies to general coverage problems beyond the present application.

Secondly, we derive demand characteristics from the ambulance response-time distribution through a survival-based weighting scheme combined with spatial covariates, so that the physiological significance of response time in cardiac-arrest survival is built into the demand model rather than approximated by proximity alone.

Lastly, and most distinctively, we formulate the facility-activation decision as a generalised Bayesian inference problem over the binary design space via a Gibbs posterior \cite{bissiri2016}. This is fundamentally different from any existing Bayesian reliability work, in which distributional assumptions are made in order to deploy Gibbs \emph{sampling} as a device for posterior inference. Here, by contrast, the Gibbs \emph{posterior} is a generalised (quasi-)Bayesian construction that dispenses with a likelihood altogether, replacing it with a structurally coherent, decision-theoretic loss and thereby avoiding the risk of model misspecification. To our knowledge, this is the first use of the Gibbs-posterior framework for reliability-informed design of a complex logistic network, and a clear theoretical stepping stone in Bayesian learning approaches for reliability analysis without any distributional assumption.

Beyond these immediate improvements, our formulation, to our knowledge, is the first to incorporate structural preferences arising from engineering constraints on cost, empirical demand, and proximity to existing \ac{ems} infrastructure into the network design of \ac{uav}-assisted \ac{aed} delivery through a logistic type prior distribution. Since the design is obtained as a posterior distribution rather than a point estimate, we collect posterior samples of the activation probabilities of candidate locations across season–hour configurations, which exposes how seasonal weather and hourly demand interacts with individual stations. For decision making purpose, we consider a site to be ineffective if the activation probability is near zero, whereas a site that is important across different conditions is considered as important and retained in the design. Furthermore, to ensure a site remains operational across seasons, we select locations from the activation probability marginalised over different seasons and time of day. Our framework also considers an empirical cost analysis of network design and a system-level robustness evaluation, in which simulated facility failures and the tails of the environmental model are propagated through the design to quantify coverage degradation, mission-level reliability under downtime and simultaneous allocation, and tail risk under low-probability but extreme conditions.

Finally, we present a detailed uncertainty decomposition that separates aleatory from epistemic contributions across the distinct domains of the problem — and, uniquely, extends epistemic uncertainty to the design itself — underscoring the value of principled uncertainty quantification at the network-design stage. By integrating probabilistic coverage modelling, generalised Bayesian learning, and reliability assessment in a single framework, our work contributes a solid systems-oriented framework for uncertainty-aware emergency drone-network design. Moreover, due to the simple yet mathematically elegant formulation, it can be adapted readily to any uncertainty-aware design of quick-response networks which showcases the applicability of our approach for different types of problems.}

\subsection{Paper Organisation}

The rest of the paper is organised as follows: we first discuss the relevant to surrogate models in \cref{sec:model} which will be used for our main modelling framework. Then we present our novel Bayesian learning approach for network design in \cref{sec:design} and methods for computing cost-effectiveness and coverage reliability from the posterior samples in \cref{sec:post:hoc}. In \cref{sec:analysis}, we illustrate our approach using two case studies focused on Scotland. Finally, we conclude our paper in \cref{sec:conc} with discussion and future works.

\section{Surrogate Models for Uncertainty Propagation}\label{sec:model}

Our goal is to achieve an uncertainty aware equity based design that accounts for the uncertainty in response time by traditional \ac{ems} as well as the reliability of the alternative \ac{uav} assisted \ac{aed} delivery. Quite obviously, the reliability of the drone based response system depends on the performance of the drone, which are highly sensitive to wind. In general, the drone performance can be adversely affected by the precipitation as well but in our studies we refrain from constructing a precipitation model, since most weather resistance drones can withstand precipitation upto 50 mm/h \citep{gao_2021_weather}, whereas the daily average in Scotland is about 1-5 mm. Similar arguments also stand for atmospheric temperature which is usually within the operating range. Therefore, we restrict ourselves with three different surrogate models that we will utilise for designing the drone network. 

\subsection{Surrogate Modelling Strategy}

Our main goal of constructing a surrogate model is to sample from the distribution of quantity of interest given a set of design. In our problem, all the parameters of interest have strictly positive support. Therefore, to avoid having negative values, we consider a log-transformed modelling framework. Such that
\begin{equation}
	y_i = f(\mathbf{D}_i) + \epsilon_i,
\end{equation}
where $\mathbf{D}_i$ denotes the input design vector, $y_i$ the log-transformed response, $f(\cdot)$ an unknown functional relationship, and $\epsilon_i\sim \mathcal{N}(0,\sigma^2)$ the random noise present in the model. Since our main objective is to propagate the uncertainty associated with parameters of interests, we consider techniques that are capable of estimating the random noise with a principled likelihood function. Therefore, we consider ordinary least squares method to check whether the data suggests a strong linearity and consider Gaussian process regression as a non-parametric modelling technique to check non-linear effects. A detailed discussion on the predictive framework for linear models and Gaussian process regression models is provided in the supplementary material.

\subsubsection{Validation Metrics}

We consider out-sample performance of the surrogate models for model selection. We check their predictive performance through $K$-fold cross-validation technique. That is, for each dataset, models are trained on $K-1$ partitions of the data and evaluated on the held-out subset. For validation purposes we consider three different metrics on the out of sample data. Firstly we use pseudo $R^2$ to understand the proportion of variance that can be explained using a model, afterwards we check RMSE and MAE to understand the deviations from observed values. The metrics are defined as follows:
\begin{align}
	R^2 = 1 - \frac{\sum (y_i-\hat{y}_i)^2}{\sum (y_i-\bar{y})^2},\qquad
	\text{RMSE} = \sqrt{\frac{1}{n}\sum (y_i-\hat{y}_i)^2},\qquad
	\text{MAE} = \frac{1}{n}\sum |y_i-\hat{y}_i|.
\end{align}
For all models across different datasets, we provide both mean and standard deviation of the metrics across $K$ folds. All the model selection tables are provided in supplementary material due to brevity of spaces.

\subsection{Ambulance Response Surrogate}\label{sec:amb:surr}

To construct an equity based design, that is to prioritise the areas which are currently underserved, we first need to evaluate the probability of covering those areas in 6 minutes by already existing \ac{ems} infrastructure. However, using raw observational data can be unreliable as performance of an emergency response vehicle can be affected by several external factors leading to different response time in future. Therefore, we build a surrogate model for estimating the ambulance response time. 

For that, we use \ac{ohca} data provided by \ac{sas}. The dataset after relevant cleaning consists of 3065 observations indicating the call location; the time of the day $(H)$; the response time $(T^A)$ in minutes; and other patient specific data. However, a major gap in the data collection is present due to the dynamic nature of \ac{ems} vehicles. In practice, highly time sensitive emergency calls are often attended by diverting a vehicle in midway of completing a mission with lower importance. Unfortunately such cases are not indicated in the data, nor the starting point of the \ac{ems} vehicle. So, we rely on the \ac{gis} data of \ac{sas} stations in Scotland and road network data of Scotland to estimate the response time with an assumption that a vehicle starts from a base location. While this leads to a higher uncertainty in the model, it is also conservative and robust in the sense that in future for responding to an \ac{ohca} case, a vehicle may not be on the road in most cases and it indeed might start from the \ac{sas} station. 

Based on the starting point assumption, we use \texttt{sfnetwork} package \citep{sfnet2024} in \texttt{R} to locate the nearest ambulance station using \citet{dijkstra_1959_a}'s algorithm for shortest path problem, where the edge weights are assigned as the time required to complete that specific segment of the path. This gives us a theoretical time $(T^{A^*})$ based on the maximum achievable speed. Moreover, to tackle with the lack of real-time traffic information in the dataset, we consider additional variables to capture the essence of traffic data. For that, we compute the number of big $(I_b)$ and medium $(I_m)$ intersections are present in the path; total amount of degrees $(\chi^A)$ a vehicle needs to turn to reach the call location; total length of the path $(l)$; weighted population density in the whole path $(\rho^{w})$; and the hour of the day $(H)$ when the cardiac arrest happens. A detailed explanation of data curation is provided in the supplementary material. This give us the following sampling distribution:
\begin{equation}
	\log\left(T^A\right) \sim \mathcal{N}\left(f^A\left(T^{A^*}, I_b, I_m,\chi^A, \rho^{w}, H, l\right), \sigma^2_A \right)
\end{equation}

We construct seven different surrogates by fitting a linear model and 6 different GP models to estimate $f^A$. For convenience we use the following notation for denoting the models: Linear model as `LM'; GP with Gaussian kernel and constant (linear) trend as `GP1' (`GP2'); GP with Mat\'ern 5/2 kernel and constant (linear) trend as `GP3' (`GP4'); and GP with product (additive) separable kernel as `GP5' (`GP6') where we consider Mat\'ern 5/2 kernel for non-periodic components and periodic kernel for periodic components. Note that the time of the day ($H$) has a periodic effect on the traffic. Therefore, for linear model and GP1 - GP4, we consider the \texttt{sine} and \texttt{cosine} component of the hour of the day. For GP5 and GP6, we use a periodic kernel on the hour of the day instead of splitting it into \texttt{sine} and \texttt{cosine} components. 

We found that both linear and Gaussian Process models improve substantially over the baseline predictor. Among all models, GP6 achieves the highest mean $R^2$ together with the lowest RMSE and MAE, indicating a modest but consistent dominance over the linear model. This suggests the presence of mild nonlinear effect in ambulance response time that is not fully captured by linear models. However, prediction with the GP model is roughly 160,000 times more expensive than the linear model and using it as a sampling distribution can be problematic during the network design phase. Instead we chose the linear model ($R^2 = 0.188$) as the surrogate for ambulance time which has a comparable performance to GP6 ($R^2 = 0.194$). The final trained linear model is presented in \cref{tab:amb:time:lm}.
\begin{table}[ht]
	\centering
	\begin{tabular}{crrrr}
		\hline
		Variable & Estimate & Std. Error & t value & Pr($>$$|$t$|$) \\ 
		\hline
		(Intercept) & 1.7654 & 0.0395 & 44.6802 & $<$1e-4 \\ 
		$T^{A^*}$ & 0.0290 & 0.0086 & 3.3715 & 0.0008 \\ 
		$I_b$ & 0.0038 & 0.0025 & 1.5127 & 0.1305 \\ 
		$I_m$ & 0.0051 & 0.0013 & 3.8861 & 0.0001 \\ 
		$\chi^A$ & -0.0019 & 0.0006 & -2.9398 & 0.0033 \\ 
		$\rho^w$ & -0.0464 & 0.0119 & -3.9138 & 0.0001 \\ 
		$l$ & 0.0289 & 0.0066 & 4.3694 & $<$1e-4 \\ 
		$\sin(\pi H/12)$ & 0.0260 & 0.0133 & 1.9553 & 0.0506 \\ 
		$\cos(\pi H/12)$ & 0.0850 & 0.0137 & 6.2118 & $<$1e-4 \\ 
		\hline
	\end{tabular}
	\caption{Summary of linear model fit for the ambulance response time}
	\label{tab:amb:time:lm}
\end{table}

\paragraph{Model diagnostics}
The fitted model achieved a residual standard error of $0.514$ on 3056 degrees of freedom, with an overall coefficient of determination $R^2 = 0.195$ (adjusted $R^2 = 0.193$). The global F-test confirms that the predictors are jointly significant $\left(F=92.35, p<2.2\times10^{-16}\right)$, indicating that the selected covariates explain a moderate proportion of variability in log response time.

\paragraph{Interpretation}
Since the response variable is log-transformed, regression coefficients can be interpreted as multiplicative effects on the ambulance response time. We notice that $T^{A^*}, I_b,I_m$ and $l$ have positive effect on the response time. This is rather obvious as larger routes with more intersections lead to longer response times. However, we notice negative effects for $\rho^w$ and $\chi^A$, which is rather interesting. This can be explained from the fact that in dense urban areas connectivity is better which reduces the response time and easy to avoid routes with heavy traffic. Overall, the coefficient structure aligns with physical intuition: travel distance and network complexity increase response time, while dense urban regions tend to improve the accessibility. However, the modest explained variance indicates that high level of uncertainty is present in the ambulance response time. This can be improved with a better data collection routine, discussing of which is beyond the scope of this paper. { Instead we have performed a sensitivity analysis on the effect of ambulance response uncertainty on the network design which we have presented in the supplementary material.}

\subsection{Wind Surrogates}\label{sec:wind:surr}

A major part of the uncertainty in network design arises from the wind variability. So to understand and incorporate the effect of wind, we build two different surrogates to model wind speeds $(\omega_v)$ in knots and wind direction from the north $(\omega_{\alpha})$. We collect the wind data from the CEDA archive for weather data \citep{office_2023_midas} that contains around 19 million observations. We first filter the raw hourly observations to remove invalid measurements by excluding wind directions outside the range $[0,360)$ and non-positive wind speeds. Afterwards, we extract the temporal attributes (month and year) from the observation timestamps, and apply a moving-average smoothing within each spatial location to reduce high-frequency variability and the dimensionality. The data were then aggregated by British National Grid \citep{epsg_2021_osgb} coordinates  $(E,N)$, month, and year. Subsequently we map these months into four meteorological seasons $(S)$ for obtaining the seasonal wind effects such that December to February is coded as `1'; March to May is coded as `2'; June to August is coded as `3'; and September to November is coded as `4'. This gives us 42 stations each having 4 seasonal measurements totalling 168 observations. { This way, we ensure that any local variability in the wind data is smoothed out as it is irrelevant for network design. Moreover, this also ensures that the wind model is stable and not overfitted.} Using this we get the following two sampling distributions:
\begin{align}
	\log(\omega_v) \sim \mathcal{N}\left(f^{\omega_v}\left(E,N,S\right), \sigma^2_{\omega_v}\right)\qquad\&\qquad
	\log(\omega_{\alpha}) \sim \mathcal{N}\left(f^{\omega_{\alpha}}\left(E,N,S\right), \sigma^2_{\omega_{\alpha}}\right).
\end{align}

Similar to our analyses with ambulance response time, we consider one linear model and 6 different GP models. Albeit, in this case the periodic kernel is chosen for the season of the year for GP5 and GP6 and \texttt{sine} and \texttt{cosine} components of the season variable are considered for the other models. For model selection we consider 10-fold cross-validation. All the model selection table is presented in supplementary material.

\subsubsection{Wind Speed}

We observed that Gaussian process models substantially outperform linear models in terms of predictive performance. Among the candidate models, GP6, which uses a separable additive kernel combining a periodic kernel for seasonal effects and a Mat\'ern $5/2$ kernel for spatial dependence, achieves the best performance ($R^2 \approx 0.94$) together with the lowest RMSE and MAE. This indicates the presence of strong nonlinear spatio–temporal dependence in wind speed.

\paragraph{Estimated kernel parameters}
For the periodic kernel, the estimated period is $P \approx 3.08$, with scaling parameter $s \approx 0.36$ and variance $\sigma^2 \approx 7.9\times10^{-3}$. This indicates a clear repeating seasonal pattern in the wind direction, with relatively rapid changes occurring within each seasonal cycle. The relatively small scaling parameter suggests that wind direction responds sensitively to changes within the seasonal variable, producing noticeable variation across the seasonal cycle.

Spatial dependence in wind direction is modelled using a Mat\'ern $5/2$ kernel applied to the easting and northing coordinates. The estimated spatial length-scale parameters were $s_{\text{E}} \approx 1.06$ for easting and $s_{\text{N}} \approx 0.51$ for northing, with variance $\sigma^2 \approx 4.29\times10^{-3}$. The larger length-scale in the easting direction indicates that wind speed varies relatively smoothly from west to east, whereas the smaller length-scale in the northing direction suggests more localized spatial variation.

Overall the pseudo $R^2$ is 0.950 based on LOO suggesting that our model is very efficient in capturing the wind speed across Scotland.

\subsubsection{Wind Direction}

Similar to wind speed, Gaussian process models outperform linear alternatives for modelling wind direction, although predictive accuracy is lower due to the inherently higher variability of directional wind patterns. The best performing model again corresponds to GP6, which employs a separable additive kernel combining periodic seasonal structure with spatial Mat\'ern dependence. This model achieves $R^2 \approx 0.49$, indicating moderate predictive capability.

Note that for sanity check we train and test our model without log transformation as well due to the circular property of the angle. We notice that the best model is obtained again for GP6 but with slightly lower explained variance ($R^2\approx0.47$). Result for cross validation is provided in the supplementary material.

\paragraph{Estimated kernel parameters for wind speed}
For the periodic kernel, the estimated period is $P \approx 0.072$, with scaling parameter $s \approx 9.69$ and variance $\sigma^2 \approx 3.38$. The relatively large scaling parameter indicates that wind speed varies smoothly across the seasonal cycle, while the comparatively large variance associated with the periodic component suggests that seasonal effects contribute substantially to the overall variability in wind speed.

Spatial dependence is modelled using a Mat\'ern $5/2$ kernel applied to the easting and northing coordinates. The estimated spatial length-scale parameters were $s_{\text{E}} \approx 0.33$ for easting and $s_{\text{N}} \approx 0.57$ for northing, with variance $\sigma^2 \approx 0.18$. These values indicate that wind speed exhibits localized spatial variability across Scotland, with slightly smoother variation along the north-south direction than along the east-west direction.

Overall the pseudo $R^2$ is 0.592 based on LOO suggesting that our model can explain decent amount of variability in the wind direction across Scotland.

\subsection{Drone Surrogates}\label{sec:drone:surr}

The final set of surrogate models is constructed using data obtained from drone flight simulator that mimics Skyports hybrid \ac{vtol} drones used for flight trials \citep{pwsadmin_2024_nhs}. That is, we consider drones that have vertical take off and landing capabilities accompanied by fixed wing capabilities for cruising efficiently \citep{ozdemir_2013_design}. The simulator specifically considers a model that can reach upto 115 km/h of cruise speed when no external wind is present and can withstand upto 10 m/s of wind. The simulator also has two different battery modules: VTOL battery of capacity of 4000 mah and cruise battery of capcaity 12000 mah. For simulation purpose, we consider following design parameters:
\begin{itemize}
	\item \textit{Distance} ($d$): The total distance a drone needs to travel from the point of departure to the destination
	
	\item \textit{Heading} ($\chi$): The direction of the drone relative to the north. 
	
	\item \textit{Cruise height} ($h_c$): Pre specified cruise height for safe horizontal movement of the drone
	
	\item \textit{Wind speed} ($\omega_{v}$): The speed of the wind encountered during the flight operation.
	
	\item \textit{Wind angle} ($\omega_{\alpha}$) The direction of the wind relative to the north. 
	
	\item \textit{Elevation change} ($\Delta_h$): The total variation in altitude that a drone covers during the flight. 
	
	\item \textit{Direction change} ($\Delta_{\chi}$): The total changes in heading in the flight trajectory. 
	
	\item \textit{Payload mass} ($m_{pl}$): The weight of the payload carried by the drone.    
\end{itemize}

We use the generated dataset to create a surrogate model that can estimate the \textit{total flight time} require by a drone to  reach a location $(l_i)$ from a drone station $(x_j)$. However, we notice that the design parameters influence the drone differently in different phases of the flights which are \textit{take-off}; \textit{cruise} and \textit{landing}. Therefore, we resort to three different models for the different phases of the flight. Similar to our earlier analysis we consider a linear model and 4 GP models for comparison. However, in these problems there are no clear periodic components present like our problem with ambulance response time and wind speed. Therefore, we refrain from using GP5 and GP6. Instead for modelling the cruise phase we consider tail wind and cross wind components of the wind speed. For all the surrogates including the battery modules, we use total 225 observations and 10-fold cross-validation for model selection.

\subsubsection{Take-Off Phase}
The first phase of a drone flight is the take-off phase. For take-off time $(T^{1})$, only three design variables are important in the construction of the surrogate models, which are the cruise height $(h_c)$; the wind speed $(\omega_{v})$; and the payload mass $(m_{pl})$. So we wish to obtain the following sampling distribution:
\begin{equation}
	\log\left(T^{1}\right) \sim \mathcal{N}\left(f^{T^1}\left(h_c,\omega_{v},m_{pl}\right), \sigma^2_{T^1}\right).
\end{equation}
We found that both linear and GP models achieve strong predictive performance and GP3 attains the highest mean $R^2$ (0.87); lowest RMSE and MAE. however LM performs at par $(R^2 =0.86)$ . Given the marginal differences across models and the computational efficiency of linear models, LM is selected to model the take-off time. This gives the following final model as presented in \cref{tab:vtol:lm}. Model selection tables can be found in supplementary material.

\paragraph{Model diagnostics}
The fitted model achieved a residual standard error of $0.067$ on 221 degrees of freedom, with an overall coefficient of determination $R^2 = 0.863$ (adjusted $R^2 = 0.861$). The global F-test confirms that the predictors are jointly significant $\left(F=463, p<2.2\times10^{-16}\right)$, indicating that the selected covariates explain a significant proportion of variability in log take-off time.

\paragraph{Interpretation}
We notice that all the covariates have positive effect on the response time. This is fairly obvious as larger payload mass or stronger wind can take longer for the drone to reach cruise height. Similarly, a higher cruise height will take longer for the drone to attain.

\subsubsection{Cruise Phase}
The second phase of a drone flight is the cruise phase. For the cruise time $(T^2)$, we consider more variables in the model, which are distance $(d)$; payload mass $(m_{pl})$; direction change $(\Delta_{\chi})$; and elevation change $(\Delta_h)$. Additionally, we compute the tail wind and cross wind in the following way:
\begin{align}
	\text{cross-wind} \coloneqq \omega_{vy} = \omega_{v}\sin(\omega_{\alpha}-\chi)\qquad\&\qquad
	\text{tail-wind} \coloneqq \omega_{vx} = \omega_{v}\cos(\omega_{\alpha}-\chi).
\end{align}
Therefore, the sampling distribution for the cruise time is given by:
\begin{equation}
	\log\left(T^2\right)\sim  \mathcal{N}\left(f^{T^2}\left(d, m_{pl}, \Delta_{\chi}, \Delta_h, \omega_{v},\omega_{\alpha}, \chi\right), \sigma^2_{T^2}\right)
	\coloneqq \mathcal{N}\left(f^{T^2}\left(d, m_{pl}, \Delta_{\chi}, \Delta_h, \omega_{vx},\omega_{vy}\right), \sigma^2_{T^2}\right).
\end{equation}
We noticed that LM achieved a very high predictive accuracy ($R^2 \approx 0.94$), whereas GP models perform poorly and in some cases show unstable behaviour. While this can again be related to the fact that GP tends to act in an interpolating manner therefore the out sample predictive performance can be extremely bad. This also shows the importance of selecting a model based on out sample metrics instead of the goodness of fit of the model. The final selected linear model is presented in \cref{tab:cruise:lm}.

\paragraph{Model diagnostics}
The fitted model achieved a residual standard error of $0.09$ on 218 degrees of freedom, with an overall coefficient of determination $R^2 = 0.951$ (adjusted $R^2 = 0.949$). The global F-test confirms that the predictors are jointly significant $\left(F=698, p<2.2\times10^{-16}\right)$, indicating that the selected covariates explain a significant proportion of variability in log cruise time.

\paragraph{Interpretation}
The cruise time model in \cref{tab:cruise:lm} indicates that flight distance $d$ is the dominant factor affecting travel duration. The coefficient for $d$ is positive and highly significant ($p<1\times10^{-4}$), confirming that longer routes proportionally increase cruise time. Directional changes $\Delta_{\chi}$ also exhibit a statistically significant positive effect ($p<1\times10^{-4}$), indicating that manoeuvring along more complex trajectories slightly increases the time required to reach the destination. Wind conditions play an important role in cruise performance. In particular, the tailwind component $\omega_{vx}$ has a strong negative and highly significant effect on cruise time ($p<1\times10^{-4}$), suggesting that favourable wind direction can reduce flight time. In contrast, the crosswind component $\omega_{vy}$ does not appear to have a statistically significant influence on cruise time which can be explained with from the fact small manoeuvres alters the sign of cross winds. 

Other design variables, such as payload mass $m_{pl}$ and elevation change $\Delta_h$, are also not statistically significant predictors of cruise time compared to distance and wind alignment.

\subsubsection{Landing Phase}
Landing phase is the third phase of the drone flight. Similar to our analysis with the take-off phase, we consider the same variables for modelling the landing time $(T^3)$ which gives us the problem of finding the following sampling distribution:
\begin{equation}
	\log\left(T^{3}\right) \sim \mathcal{N}\left(f^{T^3}\left(h_c,\omega_{v},m_{pl}\right), \sigma^2_{T^3}\right).
\end{equation}
We found that linear model again provide the best overall performance. However, GP2 and GP4 also have comparable results. Considering overall performance and computational cost, LM is selected as the landing surrogate. The final linear model is presented in \cref{tab:vtol:lm}.

\paragraph{Model diagnostics}
The fitted model achieved a residual standard error of $0.342$ on 221 degrees of freedom, with an overall coefficient of determination $R^2 = 0.46$ (adjusted $R^2 = 0.45$). The global F-test confirms that the predictors are jointly significant $\left(F=63, p<2.2\times10^{-16}\right)$, indicating that the selected covariates explain a meaningful proportion of variability in log landing time.

\paragraph{Interpretation}
We notice that cruise height and wind speed have positive effect on the response time. This is fairly obvious as higher cruise height or stronger wind can take longer for the drone to reach the ground. On the other hand having a higher payload mass will actually help the drone to reach faster to the ground and therefore having a negative effect is expected.

\paragraph{Total flight time}

Across all flight phases, linear models consistently match or outperform Gaussian Process alternatives. This indicates that flight dynamics within the explored design space are adequately predicted using linear model. Now, that we have the desired surrogates for each flight phase, we wish to find the sampling distribution for the total flight time. Note that, surrogates for take-off time, cruise time and landing time can be written as a log-normal distribution. Therefore, to obtain the sampling distribution of the total flight time, we use the moment matching method proposed by \citet{marlow1967}. Then we can write the following 
\begin{equation}
	T^D \sim \mathcal{LN}\left(f^{T^D}\left(d, h_c, m_{pl}, \Delta_{\chi}, \Delta_h, \omega_{v},\omega_{\alpha}, \chi\right), \sigma^2_{T^D}\right)
\end{equation}
such that
\begin{equation}
	\sigma^2_{T^D} = \ln\left[\frac{\sum \exp\left(2 f^{T^j}(\cdot) + \sigma^2_{T^j}\right)\left(\exp\left(\sigma^2_{T^j}\right)-1\right)}{\left(\sum\exp\left(f^{T^j}(\cdot) + \frac{\sigma^2_{T^j}}{2}\right)\right)}+1\right]
	\quad\&\quad
	f^{T^D}(\cdot) = \ln\left[\sum\exp\left(f^{T^j} + \frac{\sigma^2_{T^j}}{2}\right)\right] - \frac{\sigma^2_{T^D}}{2}.
\end{equation}

Note that for sampling the total flight time, the elevation change between candidate sites and locations are obtained using \texttt{geodata} \citep{geodata} package in \texttt{R} to account for geographic elevation. However, direction change is kept at zero as in practice it has negligible effect as we can see in \cref{tab:cruise:lm}. We refrain ourselves from additional urban elevation changes from buildings as in the context of Scotland it has almost no effect and the very few tall architectures can be bypassed easily. The cruise height and payload mass is set at 50 meters and 1.38 kilograms respectively.

\subsubsection{VTOL Battery}
The VTOL battery in the hybrid drone is used to power the quadcoptors of the drone for vertical take-off and landing. So we consider the same variables to that of take-off and landing phase surrogates to model the log transformed battery consumption $(B^v)$. This gives us the following:
\begin{equation}
	\log\left(B^v\right)\sim\mathcal{N}\left(f^{B^v}\left(h_c,\omega_{v},m_{pl}\right), \sigma^2_{B^v}\right).
\end{equation}
Findings from our analysis indicate relatively weak predictability across all models for VTOL battery consumption, suggesting a high intrinsic variability. LM and GP models with linear trend effects perform similarly suggesting the presence of a weak linearity in the model. Based on overall performance and computational aspect, we select LM as the final surrogate for the VTOL battery subsystem. We present the model in \cref{tab:vtol:lm}.

\paragraph{Model diagnostics}
The fitted model achieved a residual standard error of $0.378$ on 221 degrees of freedom, with an overall coefficient of determination $R^2 = 0.336$ (adjusted $R^2 = 0.327$). The global F-test confirms that the predictors are jointly significant $\left(F=37.25, p<2.26\times10^{-16}\right)$, indicating that the selected covariates explain a moderate proportion of variability in log VTOL battery consumption.

\paragraph{Interpretation}
We notice that cruise height and wind speed have positive effect on the battery consumption. This is fairly obvious as higher cruise height or stronger wind can take longer for the drone to reach cruise height or reach the ground. On the other hand having a higher payload mass helps the drone to reach faster to the ground and therefore having a negative effect is expected.

\subsubsection{Cruise Battery}
For the consumption of the cruise battery $(B^c)$, we consider all the variables that we used earlier to model the cruise time. Therefore we can write the distribution of the battery consumption in the following way:
\begin{equation}
	\log\left(B^c\right)\sim\mathcal{N}\left(f^{B^c}\left(d, m_{pl}, \Delta_{\chi}, \Delta_h, \omega_{v},\omega_{\alpha}, \chi\right), \sigma^2_{B^c}\right)
	\coloneqq \mathcal{N}\left(f^{B^c}\left(d, m_{pl}, \Delta_{\chi}, \Delta_h, \omega_{vx},\omega_{vy}\right), \sigma^2_{B^c}\right).
\end{equation}
Similar to our analyses with cruise time, we found that the linear model outperforms all the other model. This indicates predominantly linear dependence between mission variables and cruise battery consumption. Consequently, LM is selected as the cruise battery surrogate. We present the fitted linear model in \cref{tab:cruise:lm}.

\paragraph{Model diagnostics}
The fitted model achieved a residual standard error of $0.088$ on 218 degrees of freedom, with an overall coefficient of determination $R^2 = 0.947$ (adjusted $R^2 = 0.946$). The global F-test confirms that the predictors are jointly significant $\left(F=650, p<2.2\times10^{-16}\right)$, indicating that the selected covariates explain a significant proportion of variability in log cruise battery consumption.

\paragraph{Interpretation}
The cruise battery consumption model presented in \cref{tab:cruise:lm} shows a broadly similar pattern to that of cruise time. Distance $d$ again has a strong positive and statistically significant effect ($p<1\times10^{-4}$), indicating that longer routes results to more energy consumption during the cruise phase. Direction change $\Delta_{\chi}$ also increases battery consumption ($p=0.0004$), reflecting the additional control effort required during manoeuvring. The tailwind component $\omega_{vx}$ has a strong negative effect on battery usage ($p<1\times10^{-4}$), suggesting that favourable wind conditions reduce the propulsion effort required to move forward. 

In contrast, payload mass $m_{pl}$, and the crosswind component $\omega_{vy}$ do not show statistically significant effects on cruise battery consumption. However,  elevation change $\Delta_h$ has a minor role on the cruise battery and frequent elevation changes can lead to more battery consumption.

\begin{table}[ht]
	\centering
	\begin{tabular}{c|rrr|rrr|rrr}
		\hline
		& \multicolumn{3}{c|}{Take-off time} 
		& \multicolumn{3}{c|}{Landing time} 
		& \multicolumn{3}{c}{VTOL battery} \\
		\cline{2-4} \cline{5-7} \cline{8-10}
		Variable 
		& Est. & S.E. & $p$-val 
		& Est. & S.E. & $p$-val 
		& Est. & S.E. & $p$-val \\
		\hline
		(Intercept) 
		& 1.4495 & 0.0269 & $<$1e-4 
		& 3.8494 & 0.1373 & $<$1e-4 
		& -0.9255 & 0.1515 & $<$1e-4 \\
		
		$\omega_v$ 
		& 0.0518 & 0.0026 & $<$1e-4 
		& 0.1605 & 0.0131 & $<$1e-4 
		& 0.1422 & 0.0144 & $<$1e-4 \\
		
		$m_{pl}$ 
		& 0.3421 & 0.0313 & $<$1e-4 
		& -0.8646 & 0.1597 & $<$1e-4 
		& -0.5602 & 0.1761 & 0.0017 \\
		
		$h_c$ 
		& 0.0097 & 0.0003 & $<$1e-4 
		& 0.0080 & 0.0016 & $<$1e-4 
		& 0.0064 & 0.0018 & 0.0004 \\
		\hline
	\end{tabular}
	\caption{Summary of linear model fits for take-off time (left section); landing time (middle section); and VTOL battery consumption (right section).}
	\label{tab:vtol:lm}
\end{table}

\begin{table}[ht]
	\centering
	\begin{tabular}{c|rrr|rrr}
		\hline
		& \multicolumn{3}{c|}{Cruise time} 
		& \multicolumn{3}{c}{Cruise battery} \\
		\cline{2-4} \cline{5-7}
		Variable 
		& Est. & S.E. & $p$-val 
		& Est. & S.E. & $p$-val \\
		\hline
		(Intercept) 
		& 3.6110 & 0.0353 & $<$1e-4 
		& -1.7639 & 0.0364 & $<$1e-4 \\
		
		$d$ 
		& 0.0003 & $<$1e-4 & $<$1e-4 
		& 0.0003 & $<$1e-4 & $<$1e-4 \\
		
		$m_{pl}$ 
		& -0.0535 & 0.0401 & 0.1839 
		& 0.0225 & 0.0414 & 0.5876 \\
		
		$\Delta_{\chi}$ 
		& 0.0010 & 0.0002 & $<$1e-4 
		& 0.0007 & 0.0002 & 0.0004 \\
		
		$\Delta_h$ 
		& -0.0001 & 0.0001 & 0.3402 
		& 0.0002 & 0.0002 & 0.1025 \\
		
		$\omega_{vx}$ 
		& -0.0237 & 0.0025 & $<$1e-4 
		& -0.0234 & 0.0026 & $<$1e-4 \\
		
		$\omega_{vy}$ 
		& 0.0041 & 0.0026 & 0.1125 
		& 0.0021 & 0.0026 & 0.4306 \\
		
		\hline
	\end{tabular}
	\caption{Summary of linear model fits for cruise time (left section) and cruise battery consumption (right section).}
	\label{tab:cruise:lm}
\end{table}

\begin{figure}[ht!]
	\centering
	\includegraphics[width = 0.99\linewidth]{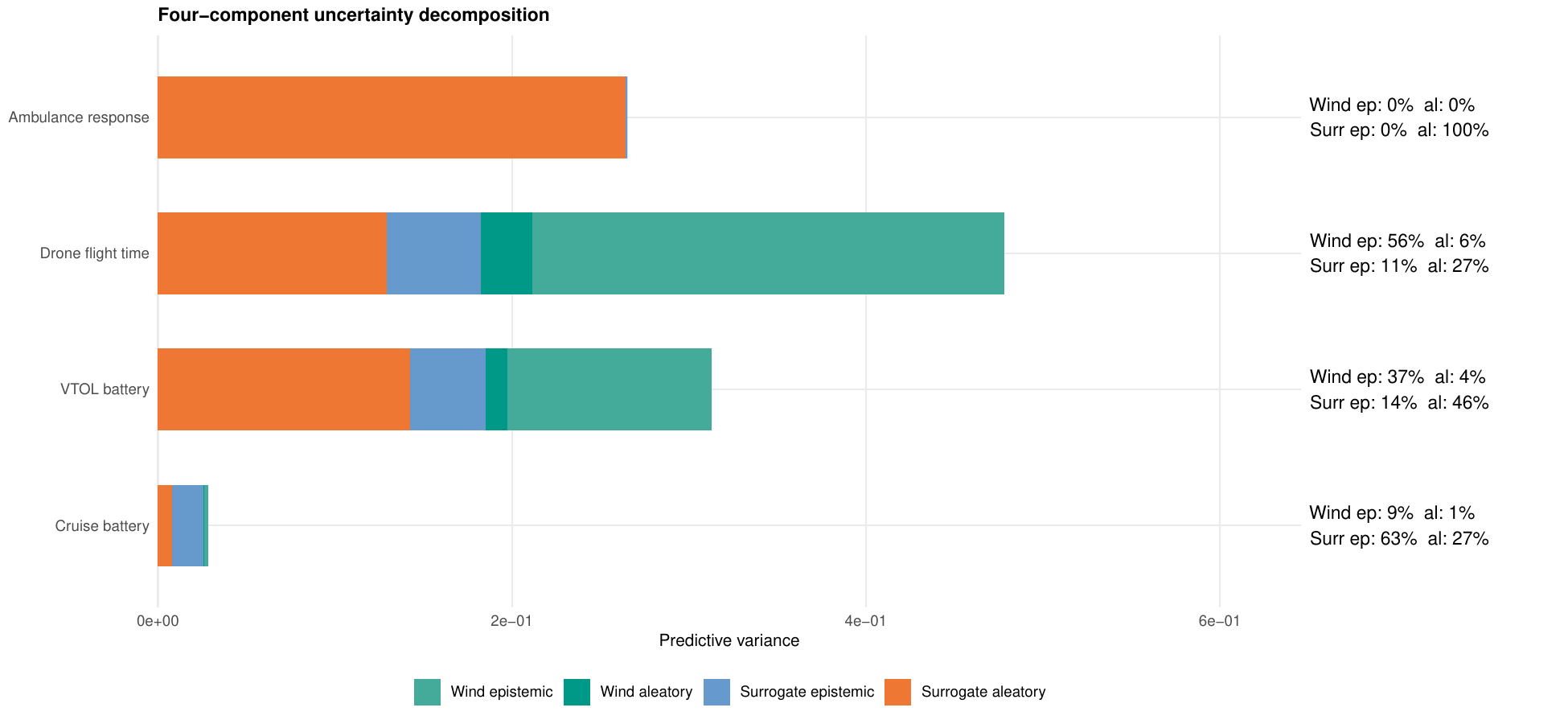}
	\caption{Aleatory and epistemic uncertainty decomposition for ambulance surrogate, wind surrogates and flight surrogates.}
	\label{fig:uq:decomp}
\end{figure}

{\subsection{Uncertainty decomposition}
	
The four-component variance decomposition in \cref{fig:uq:decomp} provides the presence of different uncertainty sources in the surrogate models. For ambulance response time, variance is almost entirely aleatory ($99.7\%$), confirming that the low explained variance ($R^2 \approx 0.19$). This suggests process irreducibility rather than model mismatch. External factors such as: traffic conditions, crew availability, and dynamic vehicle diversion leads to irreducible variability that can not be eliminated. For drone flight time, the dominant source is wind epistemic uncertainty ($55.8\%$), i.e.\ GP posterior uncertainty arising from the limited spatial points of wind observation stations; wind 	aleatory variance (nugget, $6.1\%$) and surrogate contributions ($38.1\%$) are secondary, indicating that expanding the wind observation stations would meaningfully reduce flight-time prediction uncertainty. VTOL battery consumption exhibits a different behaviour. Surrogate aleatory variance dominates ($45.6\%$), with total surrogate 	uncertainty ($59.2\%$) exceeding total wind uncertainty ($40.8\%$), suggesting that 	battery consumption during vertical flight involves process-level variability not fully 	captured by the available simulator input variables. Cruise battery is the most well behaved parameter of interest. The aleatory surrogate variance accounting for $62.8\%$ and total wind uncertainty contributing only $9.9\%$ of the uncertainty. This aligns well with the high predictive accuracy of the cruise model ($R^2 \approx 0.95$) and indicates that residual variability reflects mission-level randomness rather than wind uncertainty.
}

\section{Network Design}
\label{sec:design}

\subsection{Problem formulation}

Our objective is to find suitable locations of the drone station that forms a service coverage area. To design this coverage network, we rely on the observed \ac{ohca} locations. However, due to enormity of the problem it is practically impossible to identify these locations in a continuous spatial space. So an obvious alternative solution is to identify candidate locations where we may or may not put a drone station as part of designing the coverage network. Let $x=(x_1,\dots,x_p)$ denote the decision variable associated with selecting these candidate locations. Such that $x_j \in \{0,1\}$ and $x_j=1$ implies that the candidate location is selected for assigning a drone station. Note that the candidate locations and \ac{ohca} locations are filtered out beforehand to satisfy regulatory constraints described as no-fly zones (NFZ). 

\paragraph{Reliability component} As discussed in earlier sections, it is not reasonable to select these candidate locations using a deterministic location-allocation framework as drone coverage suffers from severe uncertainty. So, we suggest a probabilistic loss function for identifying the optimal candidate locations that improve the coverage reliability. Let $A_{ij}$ denote the probability of covering the $i$-th \ac{ohca} location in 6 minutes from the $j$-th candidate location without running out of battery. Then
\begin{equation}\label{eq:cov:prob}
	A_{ij} = P\left(T^D_{ij}\le 360\right)P\left(B^v_{ij}\le 4\right)P\left(B^c_{ij}\le 12\right)
\end{equation}
where each probability is obtained from a log-normal distribution (check \cref{sec:drone:surr} for details). { Note that, in reality the probability should not be represented as product of independent probabilities. However, our preliminary studies show that using independence indeed is conservative and reduces the computational cost as it does not require any additional simulations. We have provided the result in the supplementary material.}

Then the total aggregated probability of covering the $i$-th \ac{ohca} location in 6 minutes is given by:
\begin{equation}
	P_i(x;A)
	=
	1 - \prod_{j=1}^{p}(1 - x_j A_{ij}).
\end{equation}
From a purely probabilistic perspective $P_i(x;A)$ can find candidate locations that maximise the overall coverage. However, from a practical perspective it is desirable to select locations that also favours higher coverage of each case instead of just improving the overall coverage. So we consider a shifted logistic transformation that assigns larger penalties when coverage probability falls below 0.5. Such that,
\begin{equation}\label{eq:cov:prob:trans}
	\phi(P)
	= - a \log\left(1 + \exp[-b(P-0.5)]\right),
\end{equation}
where $b$ controls the steepness of the penalty and $a$ scales the contribution of the term. 

Now to ensure equity in network design, we introduce $e_i$ to denote the demand weights at the $i$-th call location to assess the need for placing a drone station nearby. Therefore, we formulate these demand weights such that locations with sparse \ac{ems} presence is given more importance than areas where \ac{ems} is more active. To achieve that, we consider 
\begin{equation}
	e_i \propto \left[e^{0.11T^A_i}-1\right]^{\gamma}; \qquad e_i = \frac{e_i}{\sum_g e_g},
\end{equation}
where the ambulance response time $T^A$ follows a log-normal distribution (see \cref{sec:amb:surr}). Here $e^{0.11T^A_i}-1$ acts as the odds of dying from \ac{ohca} based on current \ac{ems} response time. This is a conservative value obtained from previous empirical studies which suggest that the chance of survival of a patient is reduced by 7-10\% in every minute \citep{SINGER2020234}. The parameter $\gamma$ acts an equity parameter, which we set at 1 as it represents the linear scaling of the survival odds. Clearly for $\gamma=0$, we assign equal demand weights for all \ac{ohca} calls. This gives us the following utility related to coverage reliability:
\begin{equation}\label{eq:util:1}
	\mathcal{U}_1\left( x, T^A, \omega_{v}, \omega_{\alpha}\right)
	=
	\sum_{i=1}^{n} e_i \phi(P_i(x;A)).
\end{equation}

\paragraph{Maximal covering component} Besides coverage reliability, we also want to select drone stations that has high service potential. So, from \cref{eq:cov:prob}, we compute the 6 minutes coverage-odds utility by drones such that $\frac{A_{ij}}{1-A_{ij}}$. Using this we introduce normalised dispatch weights in the following way
\begin{equation}
	w_{ij}
	\propto \left(\frac{A_{ij}}{1-A_{ij}}\right)^{\lambda};\qquad w_{ij} = \frac{w_{ij}}{\sum_{k}w_{ik}}.
\end{equation}
Here $\lambda$ acts as a separation parameter where $\lambda\to \infty$ is equivalent to finding maximum. This way, we find the candidate location that is most efficient in serving the $i$-th demand location and ensure an overall sparser network by controlling the presence of additional stations which otherwise inflate the reliability. Then we can define the following utility that increases the average service potential of the network:
\begin{equation} \label{eq:util:2}
	\mathcal{U}_2\left( x, T^A, \omega_{v}, \omega_{\alpha}\right)
	=\frac{1}{n}\sum_{i=1}^{n}
	\sum_{j=1}^{p}
	x_j
	\, w_{ij}
	A_{ij}.
\end{equation}
From \cref{eq:util:1} and \cref{eq:util:2} and mixing parameter $\eta$, we have the following combined utility:
\begin{equation}
	\mathcal{U}\left( x, T^A, \omega_{v}, \omega_{\alpha}\right)
	\coloneqq \mathcal{U}_1\left( x, T^A, \omega_{v}, \omega_{\alpha}\right) + \eta \cdot\mathcal{U}_2\left( x, T^A, \omega_{v}, \omega_{\alpha}\right)
	=\sum_{i=1}^{n} e_i \phi(P_i(x;A)) + \frac{\eta}{n}\sum_{i=1}^{n}
	\sum_{j=1}^{p}
	x_j
	\, w_{ij}
	A_{ij}.
\end{equation}

Since both wind and ambulance operations are external random variables that are not learned in the process, we consider the following expected utility:
\begin{equation}
	\mathcal{U}( x) \coloneqq \mathbb{E}_{T^A}\left[\mathbb{E}_{\omega_{v},\omega_{\alpha}}\left[\sum_{i=1}^{n} e_i \phi(P_i(x;A)) + \frac{\eta}{n}\sum_{i=1}^{n}
	\sum_{j=1}^{p}
	x_j
	\, w_{ij}
	A_{ij}\right]\right],
\end{equation}
where $\omega_{v},\omega_{\alpha}$ follows a log-normal distribution as shown in \cref{sec:wind:surr}. Alternatively, the expected loss associated with the design can be defined as $\ell( x)\coloneqq -\mathcal{U}( x)$, which we need to minimise.

\subsection{Gibbs posterior formulation}

The loss function associated with network design leads to the following deterministic optimisation problem
\begin{equation}
	x^\ast = \arg\min_{x \in \{0,1\}^p} \ell( x),
\end{equation}
where $x^{\ast}$ is the optimal configuration for allocating drone stations. However, this optimisation problem is computationally challenging due to the combinatorial nature of the design space, which contains $2^p$ possible configurations, accompanied by the non-convex structure of the loss function. Moreover, multiple configurations may indeed exhibit similar performance in reliability, making it desirable to quantify design stability rather than reporting a single deterministic optimum.

To address these issues, we adopt a Gibbs posterior formulation over the space of decision variables. That is, we define
\begin{equation}
	\pi_\beta(x)
	\propto
	\exp\!\left\{-\beta \, \ell( x)\right\}
	\, \pi_0(x),
\end{equation}
where $\pi_0(x)$ is a subjective prior distribution arising from engineering preferences (e.g., cost, population density, spatial separation), and $\beta > 0$ is an inverse temperature parameter controlling concentration around low-loss configurations. This is equivalent to the notion of annealing as discussed by \citet{bissiri2016}. { This way the Gibbs posterior framework accommodates three distinct sources of uncertainty. The \emph{aleatory} or irreducible uncertainty present in wind conditions and ambulance response times is calculated using nuggets and residual variance. The \emph{epistemic} uncertainty is captured through the GP posterior variance and noise estimate of linear models. These two types of uncertainty are decomposed explicitly in \cref{fig:uq:decomp} and are propagated through the SAA scenarios in the expected utility (Eq.~21). The \emph{design} uncertainty, that is the uncertainty over which network configuration is most probable is quantified directly by the spread of the Gibbs posterior in Eq.~23. The design uncertainty is the distinctive contribution of our decision theoretic Bayesian formulation which allows us to examine the posterior activation probabilities that characterise design stability across the full configuration space.} Moreover, this formulation has several other advantages. Firstly, it avoids introducing a misspecified likelihood for the decision variable $x$, which often is the case as we do not know true data generating distribution. Instead, the Gibbs posterior directly incorporates the expected loss in a decision-theoretic manner. Secondly, as $\beta \to \infty$, the distribution concentrates on the global minimiser $x^\ast$, recovering the deterministic optimisation solution. 

Note that the proposed Gibbs posterior formulation is justified as a coherent update of prior beliefs under a general Bayesian framework \citep{bissiri2016}. This happens as the design space $\{0,1\}^p$ is finite and the expected reliability loss is bounded. Subsequently the normalisation constant $Z_\beta=\sum_{x\in\{0,1\}^p}\exp\{-\beta\,\ell( x)\}\,\pi_0(x)$ is finite, ensuring the Gibbs posterior is proper. 

\subsection{Prior Construction}
The prior distribution encodes practical design preferences for candidate locations. Specifically, the log-prior is defined as
\begin{equation}
	\log \pi_0(x)
	=
	\sum_{j=1}^{p}
	x_j
	\left[
	\theta_0
	+\theta_1\log(\rho_j)
	-\theta_2c_j
	+\theta_3 t_j
	\min\!\left(
	\frac{\delta_j-d_{\mathrm{thresh}}}{d_{\mathrm{thresh}}},
	0
	\right)\right]
	,
\end{equation}
where $\rho_j$ denotes the population density surrounding the $j$-th candidate location, $c_j$ represents the design cost, $t_j$ indicates whether the site corresponds to a new potential location, $\delta_j$ denotes the distance between a station to nearest existing infrastructure, and $d_{\mathrm{thresh}}$ is distance threshold.

The population density term encourages placing stations in areas with higher potential demand, while the cost term penalises expensive designs. Since existing infrastructure locations typically have lower infrastructure cost, this formulation implicitly favours the reuse of existing sites. The distance-based term is applied to avoid placing multiple drones in very close proximity to existing infrastructure. This way, it avoids putting multiple sites in urban locations where large clusters of \ac{ohca} cases are present. 

The parameters $\theta_1,\theta_2$ and $\theta_3$ act as weights for the engineering constraint variables and we set these as the inverse of the maximum of the corresponding variables. The intercept parameter $\theta_0$ acts as a sparsity control that regulates the overall number of drone stations, which we set using a prior predictive check.

\subsection{Posterior sampling scheme}

The posterior distribution over network configurations does not admit an analytically tractable expression due to the combinatorial nature of the decision variable $x \in \{0,1\}^p$. Therefore, we employ a Metropolis-Hastings (MH) sampling scheme to draw samples from the Gibbs posterior. We adopt a simple single-site flip proposal mechanism. At each iteration, a candidate site index $j \in \{1,\dots,p\}$ is selected uniformly at random and the proposed configuration $x'$ is obtained by switching the activation state of the $j$-th site. That is $x'_j = 1-x_j$, $x'_k = x_k$ for $k\neq j$. Then the MH acceptance probability of the proposed configuration becomes
\begin{equation}\label{eq:acc:prob}
	\alpha(x,x') 
	=\min\left(1, \exp\left[-\beta\left(\ell( x')-\ell( x)\right)
	+ \log \pi_0(x')-\log \pi_0(x)\right]\right).
\end{equation}
Note the loss function $\ell( x)$ requires evaluating expectations over environmental and operational uncertainties, for wind conditions and ambulance response times. To approximate these expectations efficiently, we adopt a sample-average approximation \citep{Kim2015} approach. Specifically, a fixed set of environmental scenarios is generated in advance by sampling wind speed and direction from the surrogate Gaussian process models corresponding to each season. Similarly, hourly ambulance response times are sampled from the corresponding surrogate model to construct demand weights. The expected utility is then approximated by averaging over these scenarios. This approach gives us an approximation of the expected utility within the MH iterations, which significantly reduces computational cost. We summarise the posterior sampling framework in \cref{algo:MH}.

\begin{algorithm}[h]
	\caption{Posterior sampling for drone network design}\label{algo:MH}
	\begin{algorithmic}[1]
		
		\Require Candidate sites $p$, seasons $\mathcal{S}$, hours $\mathcal{H}$, number of scenarios $K$, MH iterations $N$
		\State Initialise network configuration $x^{(0)}$
		
		\For{each season $s \in \mathcal{S}$}
		
		\State Generate $K$ environmental scenarios $\left\{\left(\omega_v^{sk}, \omega_{\alpha}^{sk}\right)\right\}_{k=1}^K$
		\State Compute coverage matrices $\left\{A_{ij}^{k}\right\}_{k=1}^K$ and dispatch weights $\left\{w_{ij}^{k}\right\}_{k=1}^K$ using sampled wind conditions
		
		\For{each hour $h \in \mathcal{H}$}
		
		\State Sample $K$ ambulance response times and construct demand weights $\left\{e_i^k\right\}_{k=1}^K$
		
		\For{$t=1,\dots,N$}
		
		\State Select site index $j \sim \text{Uniform}\{1,\dots,p\}$
		
		\State Propose configuration by toggling site $j$ such that $x'_j = 1-x_j$
		
		\State Compute expected loss using SAA
		\begin{equation*}
			\ell(x')
			=
			-\frac{1}{K}
			\sum_{k=1}^K
			\left[\sum_{i=1}^{n} e^k_i \phi\left(P_i\left(x';A^k\right)\right) + \frac{\eta}{n}\sum_{i=1}^{n}
			\sum_{j=1}^{p}
			x'_j
			\, w^k_{ij}
			A^k_{ij}\right]
		\end{equation*}
		
		\State Compute acceptance probability using \cref{eq:acc:prob} and accept proposal with probability $\alpha$
		
		\EndFor
		
		\State Store posterior samples for $(s,h,t)$
		
		\EndFor
		
		\EndFor
		
	\end{algorithmic}
\end{algorithm}

\section{Post-hoc assessments}\label{sec:post:hoc}

\subsection{Posterior Site Selection}

Let $x_{j}^{(s,h,t)}$ denote the posterior activation variable of site $j$ under season $s$ and hour $h$ at the $t$-th iteration. Using these samples we first compute the posterior activation probability for each site under each season-hour configuration
\begin{equation}
	p_{j}^{(s,h)} =
	\mathbb{E}[x_j \mid s,h] \approx \frac{1}{N}\sum_{t=1}^{N}x_j^{(s,h,t)}.
\end{equation}
Next, we obtain the seasonal activation probability by averaging across different times of the day within the same season $\hat{p}_{j}^{(s)}=\frac{1}{24}\sum_{h=1}^{24} p_{j}^{(s,h)}$. { The seasonal probabilities $\hat{p}^{(s)}_j$ can be used to understand how specific sites respond to variable wind conditions across the year. Sites with low seasonal activation probabilities in winter or autumn indicate locations where drone deployment is unreliable due to heavy winds.

For the final infrastructure decision, which remains fixed throughout the year, the yearly activation probability is obtained by averaging over all seasonal activation probabilities:
\begin{equation}
	\hat{p}_j = \frac{1}{4} \sum_{s=1}^{4} \hat{p}^{(s)}_j.
\end{equation}
Rather than applying a fixed threshold $\tau$ to $\hat{p}_j$, we select the top $k$ sites by ranking all candidates in decreasing order of $\hat{p}_j$ and retaining the $k$ highest-ranked locations, where
\begin{equation}
	k = \operatorname{median}_{s,h}
	\left\{ \operatorname{median}_t \sum_{p} x_p^{(s,h,t)} \right\}
\end{equation}
is the median posterior network size across all season-hour configurations obtained through MCMC samples of the posterior. The final site configuration is therefore
\begin{equation}
	x^\star_j = \mathbb{I}\left(
	\hat{p}_j \geq \hat{p}_{(k)}
	\right),
\end{equation}
where $\hat{p}_{(k)}$ denotes the $k$-th largest value among $\{\hat{p}_j\}$.
	
This selection rule has two advantages over a fixed threshold. First, it respects the posterior evidence on network size directly: rather than fixing a subjective choice for a threshold. We derive $k$ from the posterior distribution over the decision space $x$ and therefore the site selection can adapt to the concentration behaviour of the Gibbs posterior for the given region. Second, it guarantees a deterministic network size, avoiding the ambiguity that arises when multiple sites have activation probabilities near the threshold boundary. Using the resulting vector ${x}^\star$ we evaluate other metrics such as cost-effectiveness and coverage reliability.}

\subsection{Cost effectiveness}

\paragraph{Health outcome model}

The primary benefit of drone-assisted \ac{aed} delivery is the reduction in time to first defibrillation. In the United Kingdom, the \ac{nice} guidelines suggest that a medical intervention is considered cost effective when the cost per \ac{qaly} gained lies between £25k and £35k \citep{nice_cost_4_2022}.

Since detailed \ac{qaly} estimates specific to the UK are not available, we adopt a conservative estimation strategy based on empirical statistics reported by \citet{sas_ohca_report}. In particular, drone-delivered \acp{aed} can only benefit patients whose initial cardiac rhythm is shockable and whose collapse is witnessed by a bystander. These cases are classified as the \ac{ucg}. Approximately $22\%$ of all \ac{ohca} incidents fall into this category, and among them $26\%$ of patients survive beyond 30 days following hospital discharge. Long-term survival studies suggest that the average life expectancy of survivors is approximately 12 years \citep{Andrew1104}. 

\paragraph{Expected survival benefit}

Let $SP(T)\coloneqq e^{-0.11T}$ denote the survival probability associated with defibrillation after $T$ minutes. Let $T_i^A$ denote the ambulance response time at location $i$ and $T_{ij}^D$ denote the drone travel time from station $j$ to location $i$. Because both response times are subject to uncertainty (due to environmental and operational variability), we evaluate the expected survival benefit using $K$ stochastic simulation.

For each simulated environmental scenario $k$, the effective drone response time is given by $T_{i}^{D,(k)} = \min_{j: x_j=1}\left\{T_{ij}^{D,(k)} + 2\right\}$ where the additional 2 minutes accounts for the time to first shock following \ac{aed} arrival \citep{Smith2025}. Note that in \citet{Smith2025}, the reported drone dispatch delay includes call handling and operator processing times. Since these delays are common to both drone and ambulance dispatch, we exclude them from the comparative response-time analysis. Instead, we focus on the time required after drone arrival for a bystander to retrieve the AED and deliver the first shock. The authors reported that this process typically takes 81-126 seconds. Therefore, we conservatively assume an additional delay of 2 minutes between drone arrival and first shock. Similarly, the ambulance shock time is approximated by $T_i^{A,(k)} + 1$, where the additional minute accounts for preparation before delivering the first shock. Then the expected \ac{qaly} gain is estimated as
\begin{equation}
	\Delta \text{\ac{qaly}}
	=
	12 \times 0.22 \times 0.26
	\;
	\frac{1}{K}\sum_{k=1}^K
	\left[
	\sum_{i=1}^{n}
	\left(SP\left(T_i^{D,(k)} + 2\right)-
	SP\left(T_i^{A,(k)}+1\right)
	\right)_+
	\right],
\end{equation}
where $(\cdot)_+$ denotes the maximum between a number and zero. During this phase we also consider the drone service feasibility by sampling battery consumption. That is we check if $B_{ij}^{v,(k)}<4$ and $B_{ij}^{c,(k)}<12$ for the $k$-th scenario. If these constraints are violated, the corresponding drone response time is treated as infeasible and we set $T_{ij}^{D,(k)}$ to $\infty$. { Note that the survival probability and QALY is measured through simulated scenarios and there for also gives us the variability of QALY which we illustrate in the case studies.}

\paragraph{Cost per QALY}

In general in the design phase the cost arises from operation cost and personnel cost. Additionally, in the planning phase we want to check the expected operational cost as well to account for cost effectiveness of the network. There are different studies on the cost model of specific types of drones. In our case, for the cost analysis, we consider a rough conservative estimates provided by Skyports who conducted the flight trials. Based on this the drones and charging infrastructures cost around £50k and £10k respectively with a 20\% yearly amortisation rate. So we consider yearly cost of drones and charging ports to be $C_d = 10\text{k £}$ and $C_c = 2\text{k £}$ respectively. Besides that, we consider a conservative yearly maintenance cost of £2k ($C_m$) for each drone stations. These costs remain constant for both new and existing stations. Additionally, for new stations we will need to find a suitable drone garage or drone port facilities for which we set yearly cost to be around £5k ($C_r$). Based on these conservative estimates, the infrastructure cost associated with the network configuration $x^\star$ is
\begin{equation}
	C_{\text{infra}}
	=
	\sum_{j=1}^{p}
	x^{\star}_j (C_d + C_c + C_m + t_j*C_r),
\end{equation}
where $t_j=1$ if it's a new location and equals to zero if it's an existing \ac{sas} location. Additionally for monitoring drone operations, we would require 5 full time done operators for which we consider the salary to be around £40k per year which gives us
\begin{equation}
	C_{\text{pers}} = 200000.
\end{equation}
Lastly, for each mission we could expect upto £20 for battery charging and other miscellaneous cost. This is also a very conservative estimates as electricity in the UK costs about 26 pence per kWh, whereas the specific drone model that we consider for our analysis is roughly 0.4 kWh. Therefore the expected operational cost is given by:
\begin{equation}
	C_{\text{op}}
	=
	20 \;
	\mathbb{E}\left[
	|\mathcal{P}^{D}|
	\right],
\end{equation}
where $\mathbb{E}\left[|\mathcal{P}^{D}|\right]$ is the expected number of cases where $T_i^D + 1 < T_i^A$ and battery permits the drone to complete the mission. Finally, the expected cost effectiveness of the drone network is computed as
\begin{equation}
	\text{Cost per \ac{qaly}}
	=
	\frac{C_{\text{infra}} + C_{\text{op}} + C_{\text{pers}}}
	{\Delta \text{\ac{qaly}}}.
\end{equation}
This metric represents the expected cost required to gain one additional \ac{qaly} by utilising the drone-assisted \ac{aed} network and we use this metric to select the best network. Usually, for illustration, we consider the network configuration for which most QALY is gained and the cost per QALY is less than £30k based on \citet{nice_cost_4_2022} evaluations.

\subsection{Reliability assessment}

To evaluate the robustness of the designed drone network, we assess its performance under potential station failures. Let $x^\star$ denote the selected network configuration and let $m$ represent the number of drone stations that become unavailable due to operational disruptions such as maintenance, battery depletion, or adverse weather conditions.

Recall that the probability of successfully covering the $i$-th \ac{ohca} location from the $j$-th candidate station is denoted by $A_{ij}$ as defined in \cref{eq:cov:prob}. To model station outages, we assume that $m$ active stations fail simultaneously. Let $\mathcal{F} \subset \{j : x^\star_j = 1\}$ denote the set of failed stations such that $|\mathcal{F}| = m$. Failures are sampled without replacement with probabilities proportional to the station downtime probabilities $q_j$. The resulting effective network configuration becomes
\begin{equation}
	x^{\mathcal{F}}_j =
	\begin{cases}
		0, & j \in \mathcal{F},\\
		x^\star_j, & \text{otherwise}.
	\end{cases}
\end{equation}
Under this failure configuration, the probability that demand location $i$ remains covered is
\begin{equation}
	P_i\left(x^{\mathcal{F}};A\right)
	=
	1 - \prod_{j=1}^{p}\left(1 - x^{\mathcal{F}}_j A_{ij}\right).
\end{equation}
This way, the overall network coverage is defined as the mean coverage probability across all demand locations $C\left(x^{\mathcal{F}}\right) =\frac{1}{n}\sum_{i=1}^{n}P_i\left(x^{\mathcal{F}};A\right)$. Since the drone performance is affected by environmental conditions, the coverage probabilities $A_{ij}$ depend on uncertain wind conditions. Let $A^k$ denote the coverage matrix obtained under the $k$-th sampled wind scenario. To account for both environmental uncertainty and random station failures, we estimate the expected coverage using Monte Carlo simulations. Specifically, for $K$ wind scenarios and $R$ sampled failure sets $\mathcal{F}_r$, the expected coverage under $m$ failures is approximated as
\begin{equation}
	\widehat{C}_m(x^\star)
	=
	\frac{1}{K}
	\sum_{k=1}^{K}
	\frac{1}{R}
	\sum_{r=1}^{R}
	\frac{1}{n}
	\sum_{i=1}^{n}
	P_i\left(x^{\mathcal{F}_r}; A^k\right).
\end{equation}
Here $\widehat{C}_m(x_\star)$ represents the expected fraction of \ac{ohca} locations that remain reachable within the target response time when $m$ stations fail. By evaluating $\widehat{C}_m(x^\star)$ for increasing values of $m$, we obtain a robustness curve of the network.

{ The model above assumes that station failures occur independently as drone emergency response system is a single point failure system and it is extremely unlikely that two cases happen nearby within a short span of time. While this is a reasonable baseline, drone station availability depends on the weather condition and adverse wind conditions can affect availability of nearby stations. To assess the sensitivity of the robustness curve to such weather effects, we also consider a spatially correlated failure model based on a Gaussian copula. Stations are assigned to geographic clusters via $k$-means on site coordinates, and failure is driven by a shared cluster-level weather effect $Z_c \sim \mathcal{N}(0,1)$ combined with an idiosyncratic term $\varepsilon_j$:
\begin{equation}
	L_j = \rho\, Z_{c(j)} + \sqrt{1-\rho^2}\,\varepsilon_j 
	- (1 - \bar{A}_j),
	\label{eq:corr_fail}
\end{equation}
where $\rho \in [0,1]$ controls intra-cluster correlation and $\bar{A}_j$ is the mean coverage probability of station $j$.}

\section{Case Studies}\label{sec:analysis}

\subsection{Cardiac Arrest Statistics of Scotland}

In this section, we look into the cardiac arrest statistics of Scotland from 1st April 2022 to 31st March 2023. In this period, more than 3000 calls were recorded where a patient suffered from cardiac arrest. These calls were made predominantly from the urban areas where the majority of the population resides. As a result huge clusters of calls were formed near Glasgow and Edinburgh, the two biggest cities in Scotland. Therefore, if we want to find a drone network for emergency response then prior will tend to put more weights for drone stations near Glasgow and Edinburgh. We overcome this issue by considering different sub-regions based on \ac{nhs} Scotland's operational areas. There are 14 such sub-regions in Scotland. In our analyses, we will illustrate results for \ac{nhs} \ac{ggc} and \ac{nhs} Grampian. These test cases are particularly interesting \ac{nhs} \ac{ggc} showcases the applicability of our method for an urban setting whereas \ac{nhs} Grampian is a much bigger region that has both rural and urban areas. 

\begin{figure}[ht!]
	\centering
	\includegraphics[width = 0.99\linewidth]{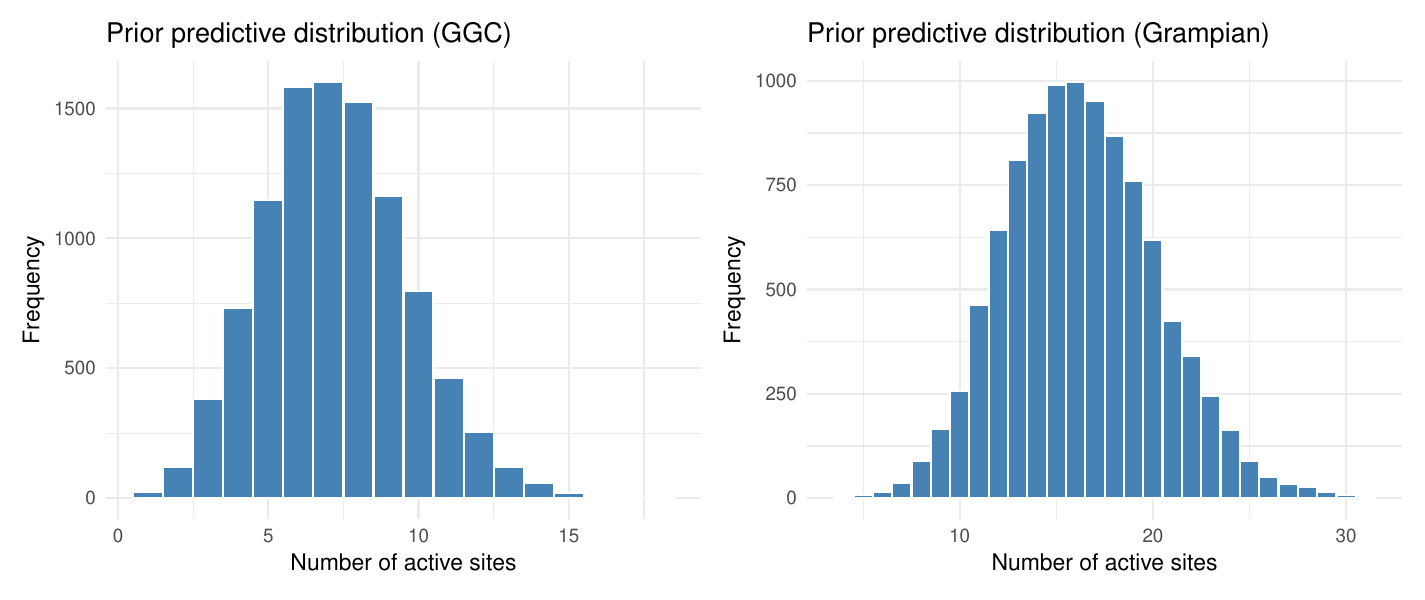}
	\caption{Prior predictive distribution on total number of active sites for NHS GGC (left) for $\theta_0=-1$ and for NHS Grampian (right) for $\theta_0=-2$.}
	\label{fig:prior}
\end{figure}

For case studies with these two cases, we consider total 100 different Monte Carlo samples for sample average approximation and 5000 MCMC iteration and 1000 burin iteration within each season-hour configuration for collecting posterior samples. Afterwards for reliability analyses we consider downtime probability to be 0.15 for new stations and 0.05 for existing stations. Additionally, for controlling the level of sparsity, we set $\theta_0$ to be -1 for NHS GGC and -2 for NHS Grampian which gives expected number of sites to be around 7 and 16 respectively. We present the prior predictive distribution below in \cref{fig:prior}.

\subsection{Greater Glasgow and Clyde}
Greater Glasgow and Clyde is \ac{nhs} Scotland's operational area in and around Glasgow. This area is the most densely populated regions of Scotland and predominantly urban in nature. This area is served by 11 \ac{sas} stations. From the 1st April 2022 to the 31st March 2023 a total 683 calls were registered as \ac{ohca}. For our analyses, we identify 50 potential locations. However, 5 potential drone stations lie inside the \ac{nfz} surrounding Glasgow airport this gives us 45 potential location out of which 9 are \ac{sas} stations. Additionally, we remove 67 \ac{ohca} locations which lie inside \ac{nfz}, as we can not serve those places using a drone.

{\subsubsection{Sensitivity Analysis and Benchmarking}

To assess the robustness of the network design, we conduct two different sensitivity analyses. First, we examine the influence of the utility hyperparameters $(a, b, \eta, \gamma)$ on the resulting network through a pilot screening sweep across 450 combinations at fixed $\beta = 15$ (provided in the supplement) with only 500 iterations. The parameters $a$ and $b$ control the scale and steepness of the penalised coverage function  $\phi(\cdot)$; $\eta$ governs the relative weight of the dispatch utility $U_2$; and $\gamma$ scales the survival-weighted demand. Across all four parameters the outputs are stable. Varying $a$ from 1 to 5 shifts the expected QALY from 93 to 94 and the number of selected stations from 13 to 15, while cost per QALY remains within the narrow range £4,612--£4,880, which is well below the NICE threshold of £25{,}000. The scale parameter $b$ shows an equally stable effect. The QALY is unchanged at 93-94 across $b \in \{5, \ldots, 25\}$ and cost per QALY spans only £4,626--£4,873. The mixing parameter $\eta$ produces the largest relative variation, with QALY rising from 92 at $\eta = 0$ (coverage reliability $\mathcal{U}_1$ only) to 95 at $\eta = 1.0$. However, the number of selected stations increases by only one site over this entire range, confirming that $\mathcal{U}_2$ refines the selection among similarly performing candidates rather than altering the size of the network. However, increase in QALY suggests that $\mathcal{U}_2$ helps in increasing the coverage unlike $\mathcal{U}_1$ which focus on the reliability aspect only. Finally, equity parameter $\gamma$ has the smallest influence of all: QALY and station count are identical at 94 and 15 respectively for all three tested values $\gamma \in \{0.0, 0.5, 1.0\}$, and cost per QALY varies by only £12. Taken together, these results confirm that the baseline values $(a=2, b=15, \eta=0.2, \gamma=1)$ adopted in the main analysis sit within will not affect the design outcome. Additionally, we run a longer analysis with 1000 iterations where $\beta$ is varied for 5 to 25 for 3 highest performing combination along with the baseline $(a=2, b=15, \eta=0.2, \gamma=1)$. We notice that all of these combinations perform similarly suggesting that our proposed approach is not very sensitive to the choice of the values of $(a, b, \eta, \gamma)$.
\begin{figure}[ht!]
	\centering
	\includegraphics[width = 0.99\linewidth]{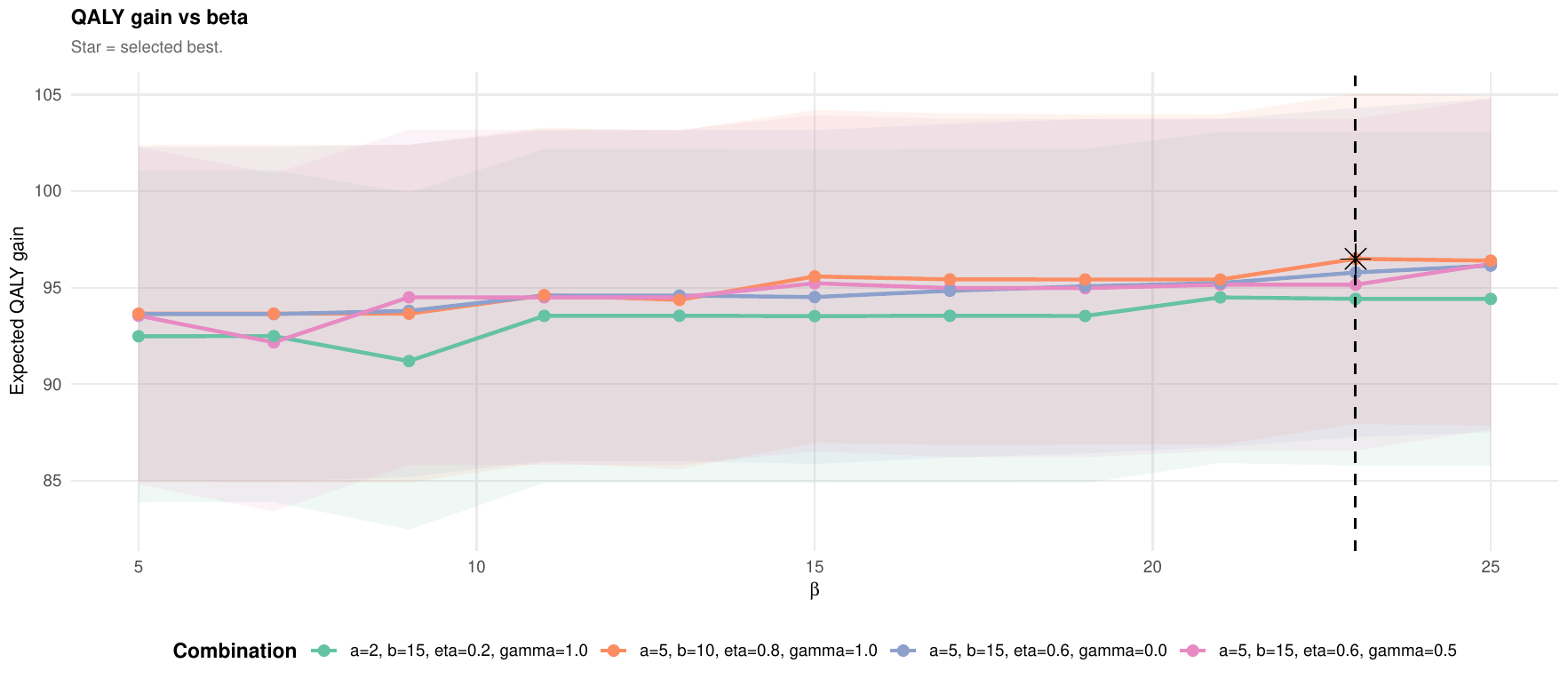}
	\caption{Longer sensitivity analysis showing QALY gain against $\beta$ for four different combinations.}
	\label{fig:results:sensti}
\end{figure}

Besides that a second sensitivity analysis is done to understand the effect of the limited explanatory power of the ambulance response surrogate ($R^2 \approx 0.19$) on the network design. The variance term $\sigma_{\text{amb}}$ is perturbed by multipliers $\{0.5\times, 1\times, 1.5\times, 2\times\}$ to simulate plausible surrogate mis-specification. The resulting site selections are compared against the baseline using Jaccard similarity of the selected sets and Spearman rank correlation of site-level activation probabilities. At half of the baseline variance ($0.5\times$) the Jaccard similarity is $0.88$ and rank correlation is $0.95$, indicating that the large majority of baseline sites are retained and the activation probability ordering is highly preserved. At twice the residual variance ($2\times$), Jaccard falls to $0.68$ and rank correlation to $0.94$. Two additional sites are exchanged relative to the baseline but the underlying activation probability of the candidate locations remains mostly unchanged. Notably, the number of selected stations is stable at 15 across all perturbation levels. However, it is worth pointing out that these analysis are done based on pilot runs of only 1000 iters. Therefore, there is high level of volatility in the network selection. Moreover, having a very strong correlation in activation probability is actually promising and suggests that chains are behaving in similar manner irrespective of the level of variance. We have summarised this results in \cref{tab:s5_amb_sensitivity}.

\begin{table}[ht]
	\centering
	\caption{Sensitivity of site selection to ambulance response surrogate
		uncertainty (NHS GGC). The residual standard deviation
		$\sigma_{\text{amb}}$ is multiplied by factors ranging from
		$0.5\times$ to $2\times$ the baseline value to simulate
		plausible surrogate mis-specification. Jaccard similarity
		measures the overlap between the selected network and the
		baseline network ($\sigma_{\text{mult}} = 1$); Spearman rank
		correlation measures the preservation of site-level activation
		probability ordering. Values close to 1 indicate robust site
		selection. The baseline $R^2 \approx 0.19$ of the ambulance
		surrogate is shown for reference.}
	\label{tab:s5_amb_sensitivity}
	\vspace{4pt}
	\begin{tabular}{cccccc}
		\toprule
		$\sigma_{\text{mult}}$ & Scenario & No.\ Sites & Baseline Sites
		& Jaccard & Rank Correlation \\
		\midrule
		$0.5\times$ & Lower uncertainty & 15 & 17 & 0.882 & 0.950 \\
		$1.0\times$ & Baseline ($R^2=0.19$) & 17 & 17 & 1.000 & 1.000 \\
		$1.5\times$ & Higher uncertainty & 15 & 17 & 0.778 & 0.965 \\
		$2.0\times$ & Higher uncertainty & 15 & 17 & 0.684 & 0.942 \\
		\bottomrule
	\end{tabular}
\end{table}

To evaluate whether the Gibbs posterior formulation is coherent with other methods and offers advantages over standard optimisation strategies, a benchmark analysis is provided against two alternatives at identical budget (to ensure valid comparison). A deterministic Maximum Coverage Location Problem (MCLP) solved via greedy forward selection is implemented using the scenario-averaged coverage matrix, and a greedy stochastic heuristic alternative that sequentially adds stations by marginal gain in expected coverage utility. \cref{tab:s9_benchmark} summarises the comparison. All three methods select the same number of stations (17) and achieve similar expected QALY gains. The Gibbs posterior yields $96.5 \pm 8.6$ QALY, MCLP yields $95.3 \pm 8.7$ QALY, and the greedy stochastic heuristic yields $92.5 \pm 8.8$ QALY. The Gibbs posterior therefore performs marginally better than MCLP by approximately 1.2 QALY and the greedy heuristic by 4.0 QALY. An interesting difference can be seen in network robustness. Coverage under one simultaneous station failure is $98.9\%$ for the Gibbs posterior, $99.6\%$ for MCLP, and $99.4\%$ for the greedy heuristic. This happens as the other two methods are not actively propagating uncertainty which therefore inflates the robustness score that can give a false sense of reliability. The Jaccard similarity between the Gibbs posterior and MCLP solutions is only $0.48$, and between the Gibbs posterior and the greedy solution only $0.36$. That is fewer than half the stations are shared between any pair of methods despite comparable aggregate performance. This indicates that the Gibbs posterior identifies a qualitatively different
spatial configuration to ensure seasonal wind variability, and survival-weighted demand equity instead of giving an approximate deterministic solution. The QALY advantage of the Gibbs posterior framework over the greedy baseline is can be seen as four additional lives saved every year within the Utstein Comperator group, which is a medically meaningful
difference that is achieved directly from the principled propagation of environmental
uncertainty and ground based response uncertainty into the decision problem.


\begin{table}[ht]
	\centering
	\caption{Benchmark comparison of the proposed Gibbs posterior against a
		deterministic Maximum Coverage Location Problem (MCLP) solved via
		greedy forward selection with 2-opt local search, and a greedy
		stochastic heuristic. All three methods are evaluated at identical
		budget ($k = 17$ stations) using the same wind and demand scenarios.
		QALY gain is reported as mean $\pm$ SD across all season-hour
		scenario combinations. Rob~$m=1$: expected coverage when one station
		fails simultaneously. Jaccard: fraction of sites in common with the
		Gibbs posterior solution ($= 1$ by definition for Gibbs).}
	\label{tab:s9_benchmark}
	\vspace{4pt}
	\begin{tabular}{lcccc}
		\toprule
		Method & No.\ Sites & QALY Gain & Rob $m=1$ & Jaccard \\
		\midrule
		Gibbs posterior (proposed) & 17 & $96.5 \pm 8.6$ & 98.9\% & 1.00 \\
		Deterministic MCLP         & 17 & $95.3 \pm 8.7$ & 99.6\% & 0.48 \\
		Greedy stochastic          & 17 & $92.5 \pm 8.8$ & 99.4\% & 0.36 \\
		\bottomrule
	\end{tabular}
\end{table}
}

\subsubsection{Results}

\begin{figure}[ht!]
	\centering
	\includegraphics[width = 0.99\linewidth]{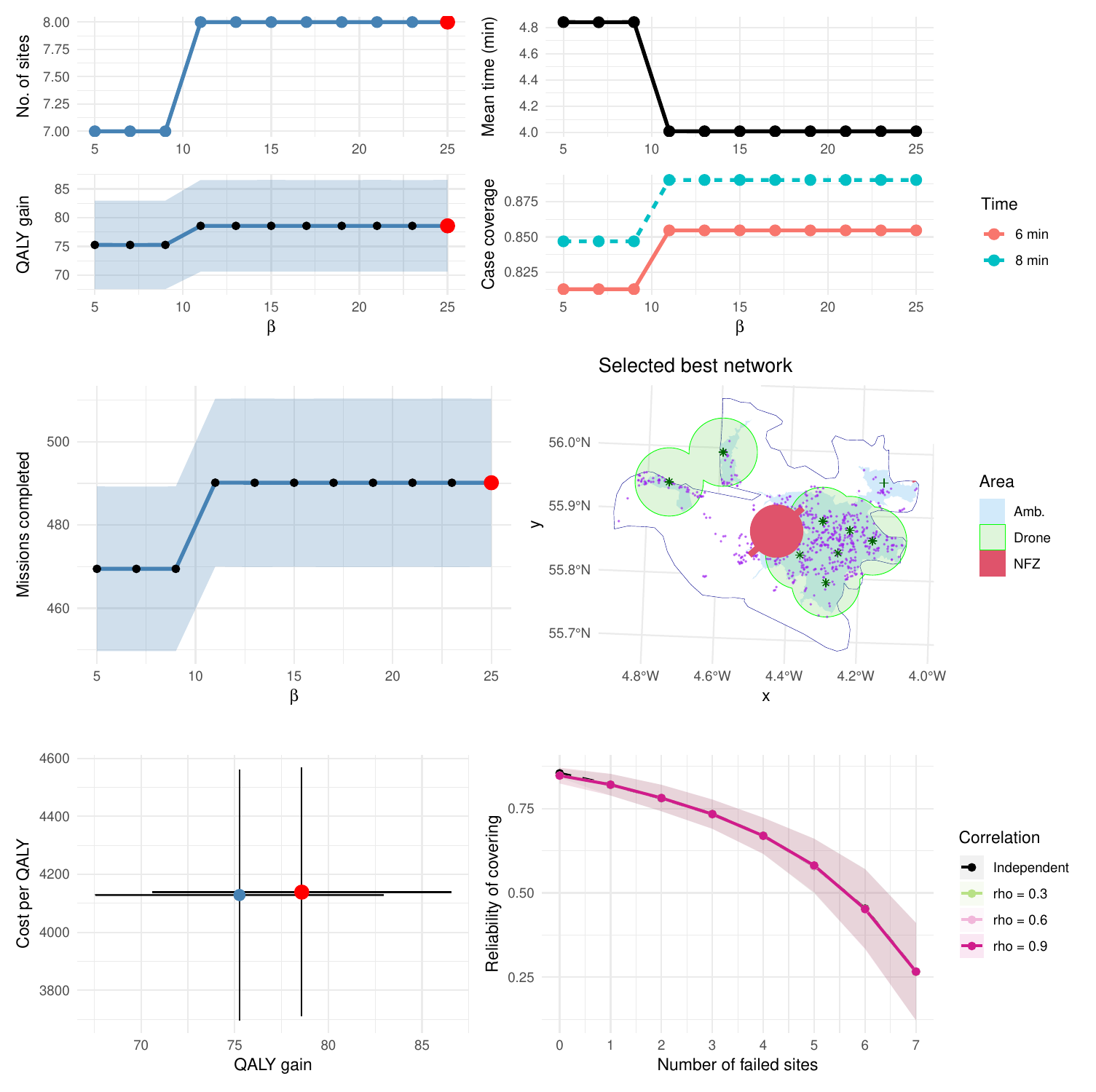}
	\caption{Summary of our analyses with NHS GGC. The upper-left panel shows number of sites and QALY gain against $\beta$; the upper-right panel shows mean drone response time and number of cases covered in 6 and 8 mins; the middle-left panel shows the number of cases where drone reaches before ambulance; the middle-right panel shows the final network design; the lower-left panel shows the cost-effectiveness of the designs; and the lower-right panel shows the coverage reliability of the designs}
	\label{fig:results:ggc}
\end{figure}

For NHS GGC we evaluate the network for $\beta$ from $5$ to $25$ with 11 equidistant grid points. For each value of $\beta$, the resulting network is assessed in terms of the number of selected sites, expected QALY gain, expected number of completed missions $\left(\mathbb{E}\left[|\mathcal{P}^{D}|\right]\right)$, cost per QALY, drone response times and robustness to station failures as summerised in \cref{fig:results:ggc}.

We notice an overall stable increasing trend in the number of station against the value of $\beta$ from the upper-left panel of \cref{fig:results:ggc}. This is an expected behaviour of Gibbs posterior formulation as higher values of $\beta$ encourages less weights on the prior which relaxes the engineering constraints. The smallest network occurs at $\beta=5,7$ and 9, where only 7 sites are selected. For the rest of the cases we get 8 sites in the network. We notice that the highest QALY gain is achieved for $\beta=25$ with an average gain of approximately $78.6$ QALY (standard deviation $8.0$). Moreover, under this configuration, the system is expected to complete approximately $490$ drone missions per year (standard deviation $20.2$). That is total cases when drone response time is at least 1 minute faster than ambulance response time. 

{ We present the cost effectiveness crosses in lower-left panel in \cref{fig:results:ggc}. Here the vertical lines express the standard deviation in cost per QALY and horizontal lines express the standard deviation in QALY gain}. We notice that all the configurations which we assess can be deemed cost effective. For $\beta=25$, the estimated cost is approximately £4,139 per QALY {(standard deviation £429)}. We present the mean response time and coverage in the upper-right panel. We notice that network obtained for $\beta=25$ can respond to all 610 calls with mean response time being 4.01 minutes. Moreover, the coverage in 6 minutes and 8 minutes are 86 and 89\% respectively which is significantly better than the coverage by ambulance based on recorded data as summarised in \cref{tab:cov:combined}. 

We present the spatial distribution of stations for the selected network ($\beta=25$)in the middle-right panel of \cref{fig:results:ggc}. In the plots we present unused ambulance stations with green `+' sign; new drone stations with green `x' sign; and drone location using existing ambulance stations with green `$\ast$' sign. For each selected drones the approximate drone coverage (radius = 6km) is shown with green filled area whereas the ambulance coverage is shown with blue filled area. We also present the \ac{nfz} around the Glasgow airport using red filled area and the \ac{ohca} calls by purple dots. We see that 8 out of 9 \ac{sas} stations are retained retained and no new station is selected. { However, it is possible to have more number of sites for values of $\beta>25$ in such cases we may get stations other than existing \ac{sas} stations.}

The reliability curves are presented in the lower-right panel of \cref{fig:results:ggc}. { For the sake of illustration, we consider total 50 scenarios for different level of correlation in the site failure. In particular, we use 0.3, 0.6 and 0.9 along with independent failure cases. We notice that these curves remains indistinguishable from each other as one would expect for single point failure model.} The curves represent the expected coverage as the number of unavailable stations increases. For the selected configuration ($\beta=25$), the expected coverage reliability is approximately $0.85$ when all stations are operational for all level of correlation. We notice that the coverage gradually decreases as stations become unavailable, falling to around $0.5$ when six stations fail.

\begin{table}[ht!]
	\centering
	\begin{tabular}{l|cccc|cccc}
		& \multicolumn{4}{c|}{GGC} & \multicolumn{4}{c}{Grampian} \\
		Response & Calls & Mean & $<$ 6 mins & $<$ 8 mins & Calls & Mean & $<$ 6 mins & $<$ 8 mins \\
		\midrule
		Recorded    
		& 683 & 8.43 & 24\% & 47\% & 280 & 8.60 & 32\% & 51\% \\
		Drone       
		& 610 (616) & 4.01 & 86\% & 89\% & 232 (248) & 4.33 & 77\% & 82\% \\
	\end{tabular}
	\caption{Coverage statistics comparing recorded \ac{sas} response times and drone-assisted response for NHS GGC and NHS Grampian for the best network design. Mean response times are reported in minutes.}
	\label{tab:cov:combined}
\end{table}

{
\subsubsection{Convergence diagnostics}

The MH acceptance rate is consistent across all $\beta$ combinations, with a mean in the range $0.33$--$0.37$ and the minimum and maximum per season-hour combination lying within $[0.31, 0.38]$ throughout. This indicates that the single-site flip proposal is efficient and the acceptance rates in this range is also typical for binary MH samplers and reflect a balance between accepting informative moves and rejecting low-probability proposals. The ESS of the log-posterior $\pi_\beta(\mathbf{x})$ has a mean between 68 and 73
across all season--hour chains, with a median consistently close to the mean (range
67--71). This confirms that the distribution of ESS is approximately symmetric and not
dominated by outliers. The minimum ESS across all chains ranges from approximately 45 to 60. For a $p$-dimensional binary variable visited by a single-site flip, an ESS of 70 from 4000 post-burn-in iterations corresponds to an integrated autocorrelation time of approximately 57 iterations. This is not unusual for a combinatorial design space of $2^p$ configurations. The ESS of the network size trace follows a similar pattern, with mean
values between 74 and 82 and minimum values above 58 in all cases, indicating that the sampler explores networks of varying size rather than collapsing to a fixed region.

Additionally, we investigate structural mixing of the chains and collapsing. We notice that none of the chain collapses for all $\beta$ combinations. That is no chain produces a constant network after the burn-in phase. The mean number of structural moves in the post-burn-in phase exceeds 1300 in all cases and increases monotonically with $\beta$, rising from approximately 1320 at $\beta = 5$ to approximately 1460 at $\beta = 25$. This suggests that sampler are not freezing at high inverse temperatures. Together, these diagnostics confirm that the Gibbs posterior sampler achieves adequate mixing across all season--hour chains and across the full range of $\beta$ values used in the case studies.

The results are summarised in a table provided in the supplementary material. A representative example of traceplot of the best network ($\beta=25$) is given in for all seasons and 12:00 pm. The traceplot confirms that satisfactory mixing of the chains.

\begin{figure}[ht!]
	\centering
	\includegraphics[width = 0.99\linewidth]{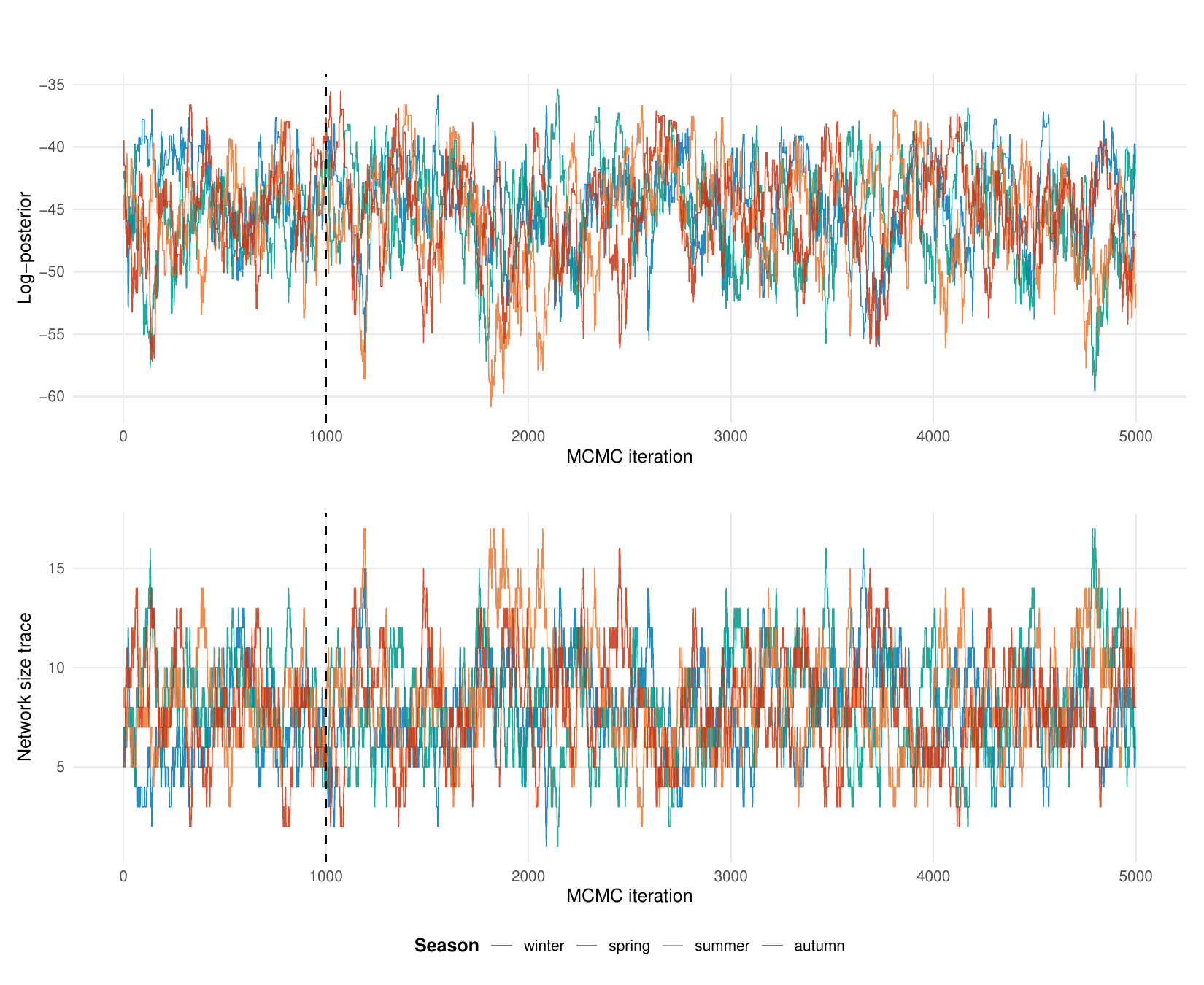}
	\caption{Traceplot of network size and log-posterior for all seasons at the afternoon}
	\label{fig:trace:grampian}
\end{figure}

}

\subsection{Grampian}
\ac{nhs} Grampian is one of the biggest boards in terms of area and has a major city which records a significant number of calls. Therefore, this board can be seen as a representation of Scotland in terms of the distribution of emergency calls in rural and urban areas. This area is served by 17 \ac{sas} stations. From the 1st April 2022 to the 31st March 2023 a total of 280 calls were registered as \ac{ohca}. In this area, we have 4 different \ac{nfz}. This reduces the potential locations to 319 out of which 17 are \ac{sas} stations. Additionally, as 32 \ac{ohca} locations lie within \ac{nfz}, the total \ac{ohca} locations reduces to 248.

\subsubsection{Results}

\begin{figure}[ht!]
	\centering
	\includegraphics[width = 0.99\linewidth]{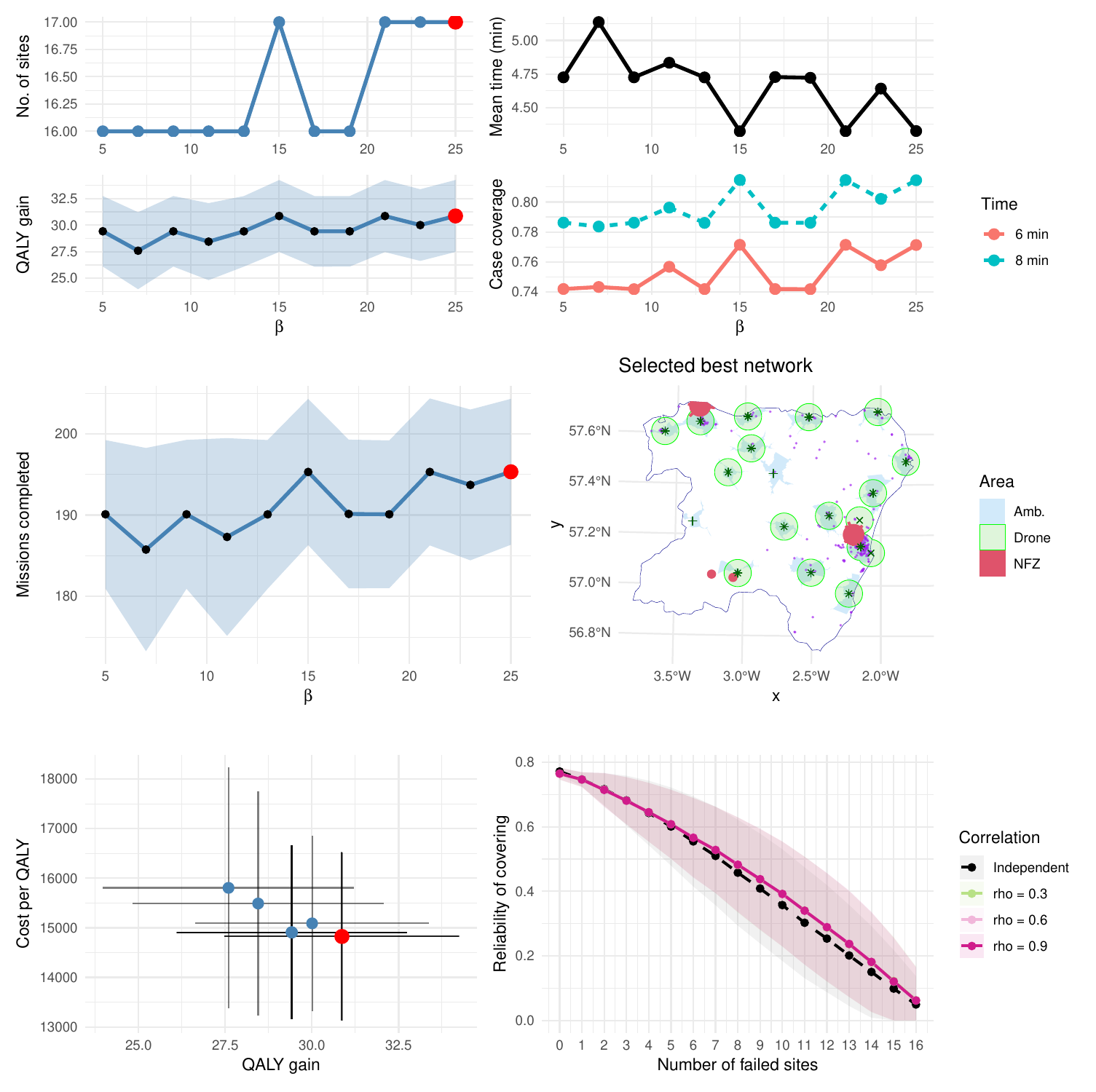}
	\caption{Summary of our analyses with NHS Grampian. The upper-left panel shows number of sites and QALY gain against $\beta$; the upper-right panel shows mean drone response time and number of cases covered in 6 and 8 mins; the middle-left panel shows the number of cases where drone reaches before ambulance; the middle-right panel shows the final network design; the lower-left panel shows the cost-effectiveness of the designs; and the lower-right panel shows the coverage reliability of the designs}
	\label{fig:results:grampian}
\end{figure}

For NHS Grampian we evaluate the network for $\beta$ from $5$ to $25$ with 11 equidistant grid points. From the upper-left panel of \cref{fig:results:grampian} we observe that the number of selected stations varies between 16 and 17 across different $\beta$ values. The largest networks are obtained for $\beta=15,21,23$ and 25 where 17 stations are selected. We observe that the highest QALY gain is achieved for $\beta = 25$, which is highlighted by the red point in the figure. This network yields an expected gain of approximately $30.9$ QALY (standard deviation $3.38$). Under this configuration, the network is expected to complete approximately $195$ drone missions (standard deviation $9.0$), which we show in middle-left panel. The cost effectiveness of different network configurations are shown in the lower-left panel of \cref{fig:results:grampian}. We notice that all configurations remain well within typical cost-effectiveness thresholds. For the selected configuration ($\beta=25$), the estimated cost is approximately £14,825 per QALY {standard deviation £1692}. We also present mean drone response times and coverage (6 and 8 minutes) in the upper-right panel. We notice that for the best network configuration the drone can respond to 232 calls out of 248 calls. The mean response time is 4.33 minutes with 77\% of all 248 calls being covered in 6 minutes, increasing it by 44\% in comparison to ambulance response which we provide in \cref{tab:cov:combined}. 

The spatial distribution of stations for the selected network ($\beta=33$) is shown in the middle-right panel of \cref{fig:results:grampian}. We observe that 15 existing infrastructure locations are retained in the optimal network while additional 2 stations are selected to extend the coverage to distant areas where ambulance response times are typically longer. We present the reliability curves in the lower-right panel of \cref{fig:results:grampian}. We notice no significant difference between the independent failure and correlated failures. This is expected as the defined network is a single point failure model. For the selected configuration ($\beta=25$), the expected coverage reliability is approximately $0.77$ when all stations are operational across all levels of correlation. 

{
Additional discussion on the MCMC convergence diagnostic is presented in the supplementary accompanied by activation probability across all seasons.}

\section{Discussion and Future works}\label{sec:conc}

In this paper we proposed a reliability-informed Bayesian framework for the design of drone-assisted AED delivery network under environmental and operational uncertainty. The proposed methodology integrates surrogate modelling, probabilistic coverage estimation, and Bayesian decision-making to build an uncertainty-aware network design for emergency response systems. By replacing deterministic coverage assumptions with response-time distributions, the framework allows environmental variability and operational uncertainty to be propagated directly into the network design process.

The study combines several modelling components within a unified framework. Surrogate models are constructed to represent ambulance response time, wind dynamics, and drone flight performance, enabling efficient uncertainty propagation during Bayesian learning. A Gibbs posterior formulation is then used to explore the space of feasible site configurations while incorporating engineering preferences through prior design structures. This probabilistic formulation allows the identification of robust network designs and provides insight into the stability of candidate station locations across seasonal and operational conditions.

We illustrate our method through case studies using geographically referenced OHCA data from Scotland. We show how we can get a robust network design in the presence of environmental variability and demand uncertainties. In addition to the network design, we perform post-hoc analysis of the system robustness under station failures and provide a cost-effectiveness analysis based on expected QALY gains. The empirical result obtained with a simulation based conservative approach shows that there is a compelling evidence on the benefits of using drone for emergency response and it can be effective for both rural and urban areas.

Certain simplifying assumptions were adopted in the present study. In particular, drone trajectories were approximated using simplified flight paths based on terrain data and regulatory constraints were incorporated at the infrastructure-selection stage rather than through detailed path-planning models. From a design perspective these are not very restrictive, as landing is the most time consuming phase in the drone flight and small manoeuvrers due to urban elevation has very little effects on the overall response time. However, it will be interesting to investigate a framework that can also perform a real-time routing analysis. Additionally, for modelling the demand factor we assume that the ambulance starts from base station. While this is a very conservative approach and gives us a robust design, it is still to consider a more dynamic traffic model that will give us a better cost-effectiveness estimates. Lastly, we would like to investigate  additional operational considerations such as in-flight failures and routing of drones from one \ac{ohca} location to another \ac{ohca} location for a more network operation level analysis.

Overall, the proposed framework provides a flexible methodology for reliability-informed planning of UAV-assisted emergency response systems. However, our Bayesian learning paradigm is flexible and can be adapted to various other sectors of network designs by integrating our method using probabilistic coverage modelling, Bayesian infrastructure selection, and system-level reliability assessment.

\section*{Acknowledgment}
This work is funded by the project CAELUS, UK Industrial Strategy
Future Flight Challenge Fund (UKRI 10023400).

\appendix

\section{Description of Surrogate Models}
	
	\subsection{Linear Models}
	
	Let $y_{i}$ denote the response variable. We fit a log-linear model such that
	\begin{equation}
		\log y_{i} = \mathbf{x}_i^\top \boldsymbol{\beta} + \varepsilon_i ,
	\end{equation}
	where $\mathbf{x}_i$ is the vector of input variables, $\boldsymbol{\beta}$ is the vector of regression coefficients, and the residual term satisfies
	\begin{equation}
		\varepsilon_i \sim \mathcal{N}(0,\sigma^{2}).
	\end{equation}
	
	\subsubsection{Sampling distribution}
	Given a new observation $\mathbf{x}_i$, the fitted model provides a predictive mean
	\[
	\hat{\mu}_i = \mathbf{x}_i^\top \hat{\boldsymbol{\beta}}
	\]
	and an associated standard error $\mathrm{SE}(\hat{\mu}_i)$ that captures uncertainty in the estimated regression mean. 	To generate predictive samples, we combine the uncertainty in the regression mean with the residual variance of the model.  The total predictive standard deviation is therefore
	\begin{equation}
		\sigma_i = \sqrt{\mathrm{SE}(\hat{\mu}_i)^2 + \sigma^{2}} .
	\end{equation}
	Then we can sample new response values for a new set of inputs from a lognormal predictive distribution
	\begin{equation}
		y_{i} \sim \text{LogNormal}(\hat{\mu}_i,\sigma_i^2).
	\end{equation}
	This way, we propagate both parameter uncertainty and residual variability when generating samples.
	
	\subsection{Gaussian Process Models}\label{apx:gp:model}
	
	Let $y$ be an output corresponding to a $p$-dimensional input $x$. We assume $y$ to be the realisation of a Gaussian random field $Y\mid X$ such that
	\begin{align}\label{eq:func:reg}
		\mathbb{E}(Y\mid X = x) = f(x)^T b\\
		\text{Cov}\left(Y\mid X=x, Y\mid X=x+h\right)  = \sigma^2 K_{\theta}(x,x+h)
	\end{align}
	where $f(x) \coloneqq \left(f_0(x), f_1(x), \cdots, f_q(x)\right)^T$ be a known vector of basis functions; $b\coloneqq (b_0,b_1, \cdots, b_q)^T$ be unknown regression coefficients; $\sigma^2$ be variance; and $K_{\theta}(x,x+h)$ is correlation kernel function with parameter $\theta$.
	
	Now, let $x^*$ be a new $p$-dimensional input. Let
	\begin{align}
		K_{\theta} &= \left[K_{\theta}(x_i,x_j)\right]_{1\le i,j\le n}\\
		K^* &= \left(K_{\theta}(x^*,x_1),\cdots,K_{\theta}(x^*,x_n)\right)^T\\
		F &= \left(f(x_1), \cdots, f(x_n)\right)^T.
	\end{align}
	Then the predicted value of $Y$ at $x^*$ and predicted variance of $Y$ at $x^*$ given by:
	\begin{align}
		\mathbb{E}\left(Y\mid X= x^*\right) =& {f^*}^T\hat{b} + {K_{\hat{\theta}}^*}^T K^{-1}_{\hat{\theta}}\left(y-F\hat{b}\right)
	\end{align}
	and
	\begin{align}
		&\text{Var}\left(Y\mid X= x^*\right) 
		= \hat{\sigma}^2\left[1 -  {K_{\hat{\theta}}^*}^T K^{-1}_{\hat{\theta}}{K_{\hat{\theta}}^*} + \left(f^* - {K_{\hat{\theta}}^*}^T K^{-1}_{\hat{\theta}}F\right)
		\left(F^TK^{-1}_{\hat{\theta}}F\right)^{-1}
		\left(f^* - {K_{\hat{\theta}}^*}^T K^{-1}_{\hat{\theta}}F\right)^T\right],
	\end{align}
	where $f^* \coloneqq \left(f_0(x^*), f_1(x^*), \cdots, f_q(x^*)\right)^T$; $\hat{b}$, $\hat{\theta}$ and $\hat{\sigma}^2$ are maximum likelihood estimates of $b$, $\theta$ and $\sigma^2$ which are solved using the following equations:
	\begin{align}
		\hat{b} &= \left(F^TK^{-1}F\right)^{-1}F^TK^{-1}y\\
		\hat{\sigma}^2 &= \frac{\left(y- F\hat{b}\right)^TK^{-1}\left(y- F\hat{b}\right)}{n - (q+1)}\\
		\hat{\theta} & = \arg\min\left[\frac{n}{2}\log \left(2\pi\hat{\sigma}^2\right)
		+\frac{1}{2}\log \left(\text{det}K\right)\right].
	\end{align}
	
	\subsection{Covariance kernel}
	In Gaussian Process (GP) regression, the covariance structure of the latent function is specified through a kernel function $K(x,x')$, which defines the correlation between function values at two input locations $x$ and $x'$. The choice of kernel determines the smoothness, periodicity, and structural properties of the resulting surrogate model.
	
	\paragraph{Periodic kernel}
	The periodic kernel is commonly used to represent repeating patterns in the data. A standard formulation is given by
	\begin{equation}
		K_{\theta}^{\text{per}}\left(x,x'\right)
		=
		\sigma^2
		\exp\left(
		-\sum_{d=1}^{D}
		\frac{2}{s_d^2}
		\sin^2\left(\pi\frac{x_d-x_d'}{p_d}\right)
		\right),
	\end{equation}
	where $\sigma^2$ denotes the variance parameter, $p_d$ denotes the period of the $d$-th input dimension, and $s_d$ is a variable-specific scaling parameter. This kernel produces functions that oscillate with period $p_d$.
	
	\paragraph{Mat\'ern $5/2$ kernel}
	The Mat\'ern family of kernels is widely used for modelling spatial or continuous inputs with controlled smoothness. The Mat\'ern $5/2$ kernel is defined as
	\begin{equation}
		K_{\theta}^{\text{Mat52}}\left(x,x'\right)
		=
		\sigma^2
		\left(1+\sqrt{5}t+\frac{5t^2}{3}\right)
		\exp(-\sqrt{5}t),
	\end{equation}
	where 
	\begin{equation}
		t =
		\sqrt{
			\sum_{d=1}^{D}
			\frac{\left(x_d-x_d'\right)^2}{s^2_d}
		},
	\end{equation}
	$s_d$ is a variable-specific scaling parameter and $\sigma^2$ is the variance parameter of the kernel. This kernel generates functions whose sample paths are twice differentiable, providing a balance between flexibility and smoothness.
	
	\paragraph{Separable kernel structures}
	When inputs consist of multiple components (for example spatial and temporal variables), composite kernels can be constructed by combining kernels defined on individual input dimensions.
	
	A \emph{separable additive kernel} is defined as
	\begin{equation}
		K^{\text{add}}_{\theta}\left(x,x'\right) =
		\sum_{d=1}^{D} K_{\theta_d}\left(x_d,x'_d\right),
	\end{equation}
	where $x_d$ denotes the $d$-th component of the input vector. This structure assumes that the effects of different input dimensions contribute independently to the covariance.
	
	Alternatively, a \emph{separable product kernel} is defined as
	\begin{equation}
		K^{\text{prod}}_{\theta}\left(x,x'\right) =
		\prod_{d=1}^{D} K_{\theta_d}\left(x_d,x'_d\right),
	\end{equation}
	which captures interactions between input dimensions by multiplying their covariance contributions. Product kernels allow correlations in one dimension to depend on the values of other dimensions.
	
	For our analyses with Gaussian process, we use \texttt{GauPro} \citep{gaupro} package in \texttt{R}.
	
	\section{Data Preparation: Ambulance Network Features}\label{app:amb:data:prep}
	
	We obtain our data from Open street map website \url{https://www.openstreetmap.org/} and prepare our data using \texttt{sfnetwork}\citep{sfnet2024} package in \texttt{R}.
	
	\subsection{Spatial Network Construction}
	
	The transportation network was from a reduced road network data to ensure any pedestrian only type paths are removed. Then we convert it into an undirected spatial network representation. Edges were subdivided to ensure accurate spatial resolution, and edge weights were defined using physical length of the road segments in meters:
	\begin{equation}
		\texttt{weight} = \text{edge length}.
	\end{equation}
	The travel time (in seconds) along each edge was computed as:
	\begin{equation}
		\texttt{time} = \frac{\texttt{weight}}{\texttt{max speed}},
	\end{equation}
	yielding a time-weighted network suitable for routing analysis. To improve routing accuracy, both emergency incident locations (``sites'') and ambulance station locations (``facilities'') were blended into the network nodes. 
	
	\subsection{Shortest-Path Cost Matrix}
	
	A travel-time cost matrix was computed between all emergency sites and all ambulance facilities using network-based shortest paths:
	\begin{equation}
		C_{ij} = \text{minimum travel time from site } i \text{ to facility } j.
	\end{equation}
	For each site $i$, the closest facility was identified as $j^* = \arg\min_j C_{ij}$. Then the corresponding minimum travel time (in minutes) is defined as $\texttt{comp\_time}_i = \frac{\min_j C_{ij}}{60}$. Sites with infinite travel time (disconnected components) were excluded from further analysis.
	
	\subsection{Path-Level Feature Extraction}
	
	For each emergency site--facility pair, the time-optimal path was extracted. From this path, the following structural and environmental features were computed:
	
	\paragraph{Intersection Counts}
	
	To get intersection counts we first compute the node-level degree centrality
	\begin{equation}
		\texttt{deg} = \text{number of incident edges}.
	\end{equation}
	Using this we compute,
	\begin{equation}
		\texttt{big\_intsects} = \sum \mathbb{I}(\texttt{deg} > 3),
	\end{equation}
	and
	\begin{equation}
		\texttt{mid\_intsects} = \sum \mathbb{I}(\texttt{deg} = 3).
	\end{equation}
	
	\paragraph{Turning Intensity}
	
	Edge azimuth (in radian) values were used to approximate cumulative turning magnitude along the route:
	\begin{equation}
		\texttt{turns} = \sum |\texttt{edge azimuth}|.
	\end{equation}
	
	\paragraph{Effective Population Density}
	
	A length-weighted population density was computed along each segment of the path (in meters):
	\begin{equation}
		\texttt{pop\_dense} =
		\frac{\sum (\texttt{density}_k \cdot \texttt{length}_k)}
		{\sum \texttt{length}_k}.
	\end{equation}
	Further more population density (per sq m) values were multiplied by a factor of 1000 for scaling purpose.
	
	\paragraph{Route Length}
	
	Total path length was calculated as:
	\begin{equation}
		\texttt{length} = \sum \texttt{edge length}
	\end{equation}
	and converted to kilometers.
	
	\subsection{Temporal Feature}
	
	Hour-of-day was extracted from the emergency incident dataset provided by Scottish Ambulance Service.
	
	\subsection{Final Modeling Dataset}
	
	The final dataset for modelling ambulance response time contains one row per emergency incident and each column indicates the following:
	
	\begin{itemize}
		\item \texttt{rec\_time}: recorded response time (minutes)
		\item \texttt{comp\_time}: computed network travel time (minutes)
		\item \texttt{big\_intsects}: number of big (more than 3) intersections (integer)
		\item \texttt{mid\_intsects}: number of 3-way intersections (integer)
		\item \texttt{turns}: cumulative turning magnitude (radian)
		\item \texttt{pop\_dense}: length-weighted population density (per 1000 sq meters)
		\item \texttt{hour}: hour of day (integer)
		\item \texttt{length}: route length (km).
	\end{itemize}
	
	This way we integrate spatial network structure, demographic context, and temporal information for subsequent modelling of the ambulance response time.
	
	\section{Model Selection Tables for Surrogates}
	\subsection{Ambulance Response Time}
\begin{figure}[ht!]
		\centering
		\includegraphics[width = 0.99\linewidth]{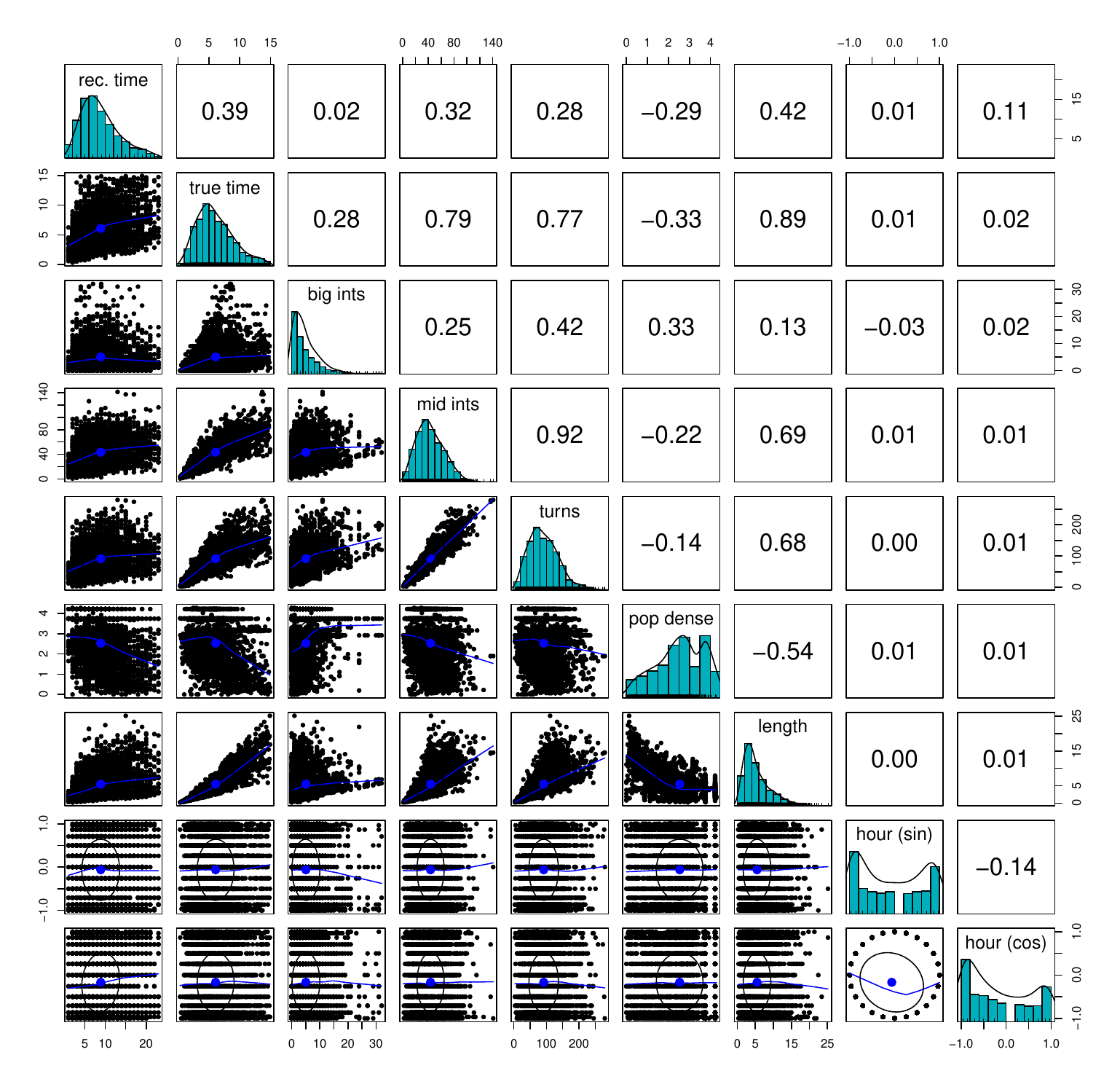}
		\caption{Scatterplot and correlation plot for the ambulance response time dataset.}
		\label{fig:corr:amb}
	\end{figure}

	We present a scatterplot diagram for the ambulance response time in \cref{fig:corr:amb}. We notice that moderate to high correlation is present between the inputs and the response variable. However, explained variance ($R^2$) remains low as many inputs are also highly correlated amongst themselves. We present the model selection table for the ambulance response time in \cref{tab:cv_ambulance}.
	
	\begin{table}[ht]
		\centering
		\caption{Three-fold cross-validation results for log-transformed ambulance response time models.}
		\label{tab:cv_ambulance}
		\begin{tabular}{lrrrrrr}
			\toprule
			Model 
			& {$R^2$ Mean} 
			& {$R^2$ SD} 
			& {RMSE Mean} 
			& {RMSE SD} 
			& {MAE Mean} 
			& {MAE SD} \\
			\midrule
			Baseline & -0.000 & 0.000 & 0.572 & 0.014 & 0.450 & 0.012 \\
			LM      &  0.188 & 0.022 & 0.515 & 0.014 & 0.396 & 0.013 \\
			GP1      &  0.086 & 0.111 & 0.546 & 0.023 & 0.424 & 0.021 \\
			GP2      &  0.016 & 0.090 & 0.567 & 0.024 & 0.436 & 0.021 \\
			GP3      &  0.080 & 0.008 & 0.548 & 0.011 & 0.426 & 0.012 \\
			GP4      & -0.102 & 0.024 & 0.600 & 0.013 & 0.463 & 0.009 \\
			GP5      &  0.043 & 0.020 & 0.560 & 0.019 & 0.434 & 0.015 \\
			GP6      &  \textbf{0.194} & 0.017 & \textbf{0.513} & 0.017 & \textbf{0.396} & 0.016 \\
			\bottomrule
		\end{tabular}
	\end{table}
	
	\subsection{Wind Models}
	
\begin{figure}[ht!]
		\centering
		\includegraphics[width = 0.99\linewidth]{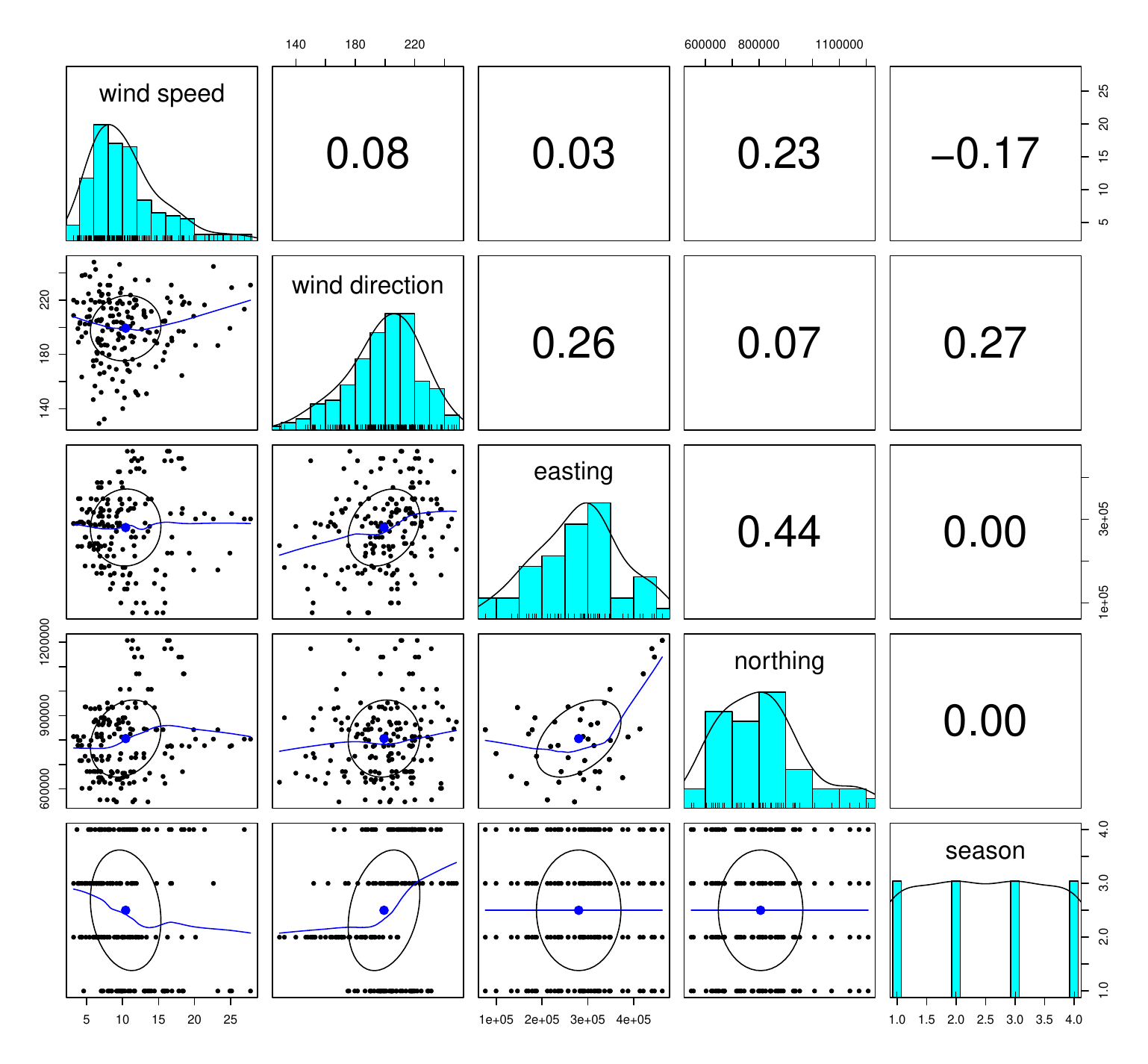}
		\caption{Scatterplot and correlation plot for wind models.}
		\label{fig:corr:wind}
	\end{figure}

	We present a scatterplot diagram for the wind speed and wind direction in \cref{fig:corr:wind}. We notice that moderate correlation is present between the inputs and the response variables. However, the there's a strong non-linear relation is present within the response variables and inputs. As a result Gaussian process models perform well very well for wind speed which we present in \cref{tab:cv_speed}. On the other hand for wind direction we see even though one of the Gaussian process performs really well, others show unstable behaviour from \cref{tab:cv_direction}.
	
	Three additional results for cross-validation for wind models is also presented. Results for wind-direction is presented in \cref{tab:cv_direction:wol} where wind direction is considered as the response variable without any log-transformation. Besides that, cross validation table with circular treatment is presented in \cref{tab:cv_vsin} and \cref{tab:cv_vcos} using LM1 and GP6.
	
	\begin{table}[ht]
		\centering
		\caption{Ten-fold cross-validation results for log wind speed models.}
		\label{tab:cv_speed}
		\begin{tabular}{lrrrrrr}
			\toprule
			Model & {$R^2$ Mean} & {$R^2$ SD} & {RMSE Mean} & {RMSE SD} & {MAE Mean} & {MAE SD} \\
			\midrule
			Baseline & -0.055 & 0.048 & 0.451 & 0.044 & 0.365 & 0.035 \\ 
			LM1 & 0.154 & 0.136 & 0.404 & 0.062 & 0.325 & 0.040 \\ 
			GP1 & 0.856 & 0.118 & 0.157 & 0.049 & 0.127 & 0.045 \\ 
			GP2 & 0.275 & 0.160 & 0.371 & 0.043 & 0.309 & 0.045 \\ 
			GP3 & 0.883 & 0.043 & 0.147 & \textbf{0.019} & 0.118 & 0.018 \\ 
			GP4 & 0.715 & 0.078 & 0.232 & 0.030 & 0.189 & 0.031 \\ 
			GP5 & 0.853 & 0.054 & 0.165 & 0.036 & 0.134 & 0.035 \\ 
			GP6 & \textbf{0.944} & \textbf{0.031} & \textbf{0.100} & 0.022 & \textbf{0.079} & \textbf{0.017} \\ 
			\bottomrule
		\end{tabular}
	\end{table}
	
	\begin{table}[ht]
		\centering
		\caption{Ten-fold cross-validation results for log wind direction models.}
		\label{tab:cv_direction}
		\begin{tabular}{lrrrrrr}
			\toprule
			Model & {$R^2$ Mean} & {$R^2$ SD} & {RMSE Mean} & {RMSE SD} & {MAE Mean} & {MAE SD} \\
			\midrule
			Baseline & -0.153 & 0.349 & 0.127 & 0.021 & 0.100 & 0.015 \\ 
			LM1 & 0.282 & 0.247 & 0.099 & 0.014 & 0.077 & \textbf{0.011} \\ 
			GP1 & -0.055 & 0.244 & 0.121 & 0.017 & 0.096 & \textbf{0.011} \\ 
			GP2 & -3.347 & 2.744 & 0.227 & 0.052 & 0.177 & 0.033 \\ 
			GP3 & -0.069 & 0.240 & 0.122 & 0.017 & 0.097 & 0.012 \\ 
			GP4 & -5.175 & 4.268 & 0.269 & 0.079 & 0.211 & 0.059 \\ 
			GP5 & -0.117 & 0.371 & 0.124 & 0.019 & 0.098 & 0.013 \\ 
			GP6 & \textbf{0.492} & \textbf{0.227} & \textbf{0.082} & \textbf{0.011} & \textbf{0.066} & \textbf{0.011} \\ 
			\bottomrule
		\end{tabular}
	\end{table}
	
	\begin{table}[ht]
		\centering
		\caption{Ten-fold cross-validation results for wind direction models.}
		\label{tab:cv_direction:wol}
		\begin{tabular}{lrrrrrr}
			\toprule
			Model & {$R^2$ Mean} & {$R^2$ SD} & {RMSE Mean} & {RMSE SD} & {MAE Mean} & {MAE SD} \\
			\midrule
			Baseline & -0.144 & 0.321 & 24.056 & 3.383 & 19.274 & 3.013 \\ 
			LM1 & 0.290 & 0.241 & 18.792 & 2.512 & 14.717 & 2.304 \\ 
			GP1 & -0.133 & 0.351 & 23.867 & 3.271 & 19.153 & 3.267 \\ 
			GP2 & -1.700 & 0.854 & 36.473 & 4.268 & 30.022 & 3.278 \\ 
			GP3 & -0.117 & 0.353 & 23.723 & 3.488 & 19.083 & 3.355 \\ 
			GP4 & -2.229 & 1.296 & 39.438 & 6.233 & 31.300 & 4.798 \\ 
			GP5 & -0.481 & 0.719 & 26.768 & 5.234 & 21.328 & 4.339 \\ 
			GP6 & \textbf{0.467} & \textbf{0.209} & \textbf{16.208} & \textbf{2.336} & \textbf{12.931} & \textbf{2.254} \\ 
			\bottomrule
		\end{tabular}
	\end{table}
	
	\begin{table}[ht]
		\centering
		\caption{Ten-fold cross-validation results for sine component of the wind through circular treatment.}
		\label{tab:cv_vsin}
		\begin{tabular}{lrrrr}
			\toprule
			Model & $R^2$ Mean & $R^2$ SD & RMSE Mean & RMSE SD \\ 
			\midrule
			LM1 & 0.28 & 0.22 & 3.60 & 1.33 \\ 
			GP6 & -1.60 & 3.64 & 5.76 & 2.80 \\ 
			\bottomrule
		\end{tabular}
	\end{table}
	
	\begin{table}[ht]
		\centering
		\caption{Ten-fold cross-validation results for cosine component of the wind through circular treatment.}
		\label{tab:cv_vcos}
		\begin{tabular}{lrrrr}
			\toprule
			Model & $R^2$ Mean & $R^2$ SD & RMSE Mean & RMSE SD \\ 
			\midrule
			LM1 & 0.17 & 0.15 & 3.84 & 0.67 \\ 
			GP6 & 0.57 & 0.09 & 2.78 & 0.51 \\ 
			\bottomrule
		\end{tabular}
	\end{table}
	
	\subsection{Drone Models}
\begin{figure}[ht!]
		\centering
		\includegraphics[width = 0.99\linewidth]{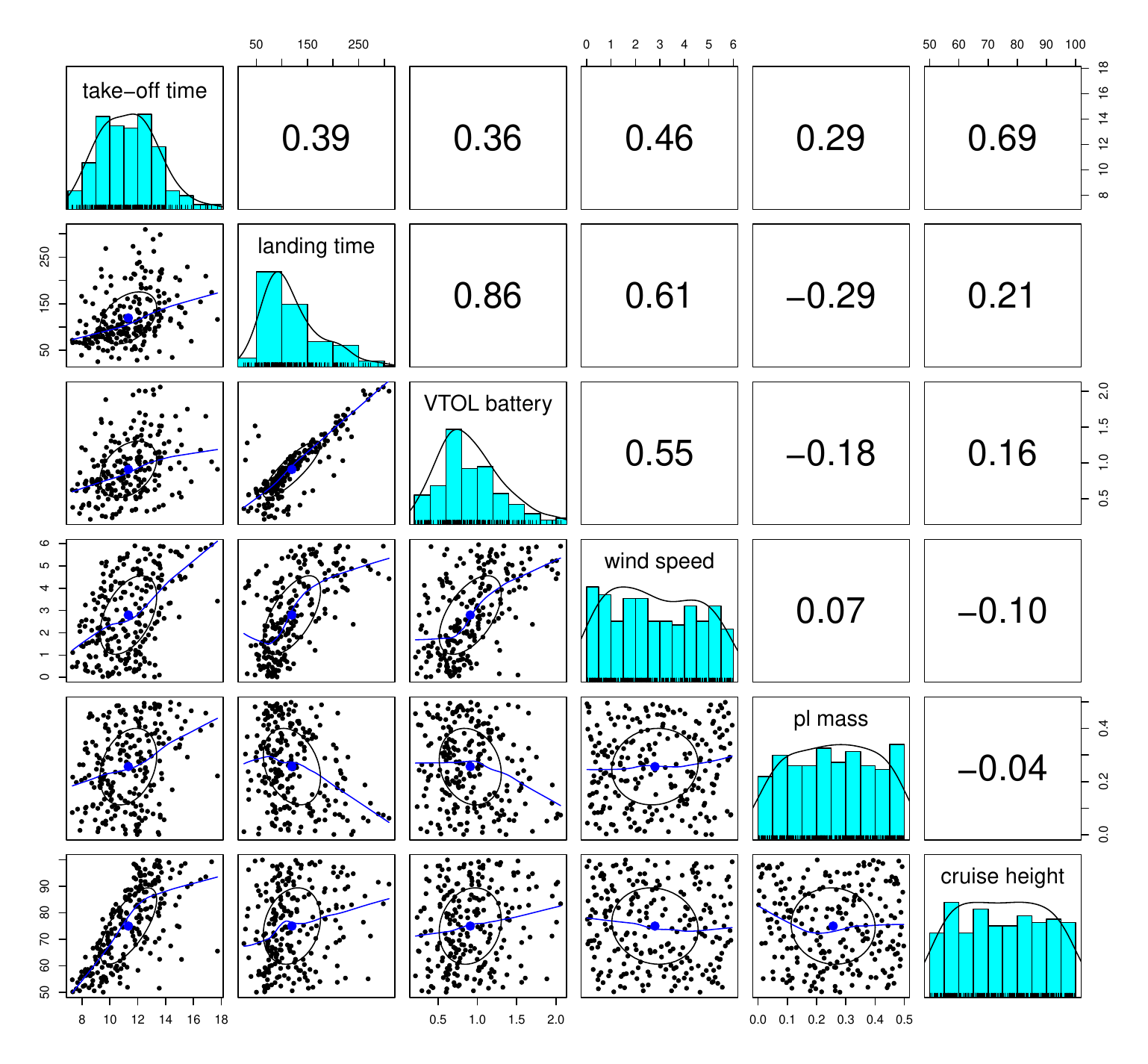}
		\caption{Scatterplot and correlation plot for VTOL models.}
		\label{fig:corr:VTOL}
	\end{figure}

	We present a scatterplot diagram for the take-off time, landing time and VTOL battery consumption in \cref{fig:corr:VTOL}. We notice that moderate correlation is present between the inputs and the response variables. We present the model selection for take off time in \cref{tab:cv_takeoff}; landing time in \cref{tab:cv_land}; and VTOL battery consumption in \cref{tab:cv_batt_v}.
	
	\begin{table}[ht]
		\centering
		\caption{Ten-fold cross-validation results for take-off time surrogate models.}
		\label{tab:cv_takeoff}
		\begin{tabular}{lrrrrrr}
			\toprule
			Model 
			& {$R^2$ Mean} 
			& {$R^2$ SD} 
			& {RMSE Mean} 
			& {RMSE SD} 
			& {MAE Mean} 
			& {MAE SD} \\
			\midrule
			Baseline & -0.045 & 0.041 & 0.179 & 0.021 & 0.149 & 0.018 \\ 
			LM1 & 0.859 & \textbf{0.130} & 0.061 & \textbf{0.030} & 0.042 & 0.011 \\ 
			GP1 & 0.819 & 0.184 & 0.066 & 0.040 & 0.047 & 0.027 \\ 
			GP2 & 0.819 & 0.131 & 0.072 & 0.029 & 0.045 & 0.012 \\ 
			GP3 & \textbf{0.866} & \textbf{0.130} & \textbf{0.060} & 0.032 & \textbf{0.041} & 0.016 \\ 
			GP4 & 0.818 & 0.137 & 0.072 & 0.031 & 0.047 & \textbf{0.010} \\
			\bottomrule
		\end{tabular}
	\end{table}
	
	\begin{table}[ht]
		\centering
		\caption{Ten-fold cross-validation results for landing time surrogate models.}
		\label{tab:cv_land}
		\begin{tabular}{lrrrrrr}
			\toprule
			Model 
			& {$R^2$ Mean} 
			& {$R^2$ SD} 
			& {RMSE Mean} 
			& {RMSE SD} 
			& {MAE Mean} 
			& {MAE SD} \\
			\midrule
			Baseline & -0.065 & 0.049 & 0.458 & 0.088 & 0.368 & 0.065 \\ 
			LM1 & \textbf{0.468} & \textbf{0.242} & \textbf{0.324} & 0.129 & \textbf{0.228} & 0.068 \\ 
			GP1 & 0.129 & 0.290 & 0.412 & 0.124 & 0.301 & 0.070 \\ 
			GP2 & 0.455 & 0.255 & 0.327 & 0.134 & 0.233 & 0.077 \\ 
			GP3 & 0.224 & 0.288 & 0.385 & \textbf{0.117} & 0.275 & \textbf{0.064} \\ 
			GP4 & 0.416 & 0.250 & 0.339 & 0.130 & 0.240 & 0.068 \\ 
			\bottomrule
		\end{tabular}
	\end{table}
	
	\begin{table}[ht]
		\centering
		\caption{Ten-fold cross-validation results for VTOL battery surrogate models.}
		\label{tab:cv_batt_v}
		\begin{tabular}{lrrrrrr}
			\toprule
			Model 
			& {$R^2$ Mean} 
			& {$R^2$ SD} 
			& {RMSE Mean} 
			& {RMSE SD} 
			& {MAE Mean} 
			& {MAE SD} \\
			\midrule
			Baseline & -0.032 & 0.028 & 0.456 & 0.069 & 0.367 & 0.060 \\ 
			LM1 & 0.288 & 0.135 & 0.376 & 0.059 & 0.291 & 0.044 \\ 
			GP1 & -0.037 & \textbf{0.108} & 0.456 & 0.066 & 0.359 & 0.053 \\ 
			GP2 & \textbf{0.294} & 0.138 & \textbf{0.374} & 0.059 & \textbf{0.290} & 0.045 \\ 
			GP3 & 0.006 & 0.281 & 0.441 & 0.073 & 0.344 & 0.054 \\ 
			GP4 & 0.281 & 0.139 & 0.377 & \textbf{0.058} & 0.291 & \textbf{0.041} \\  
			\bottomrule
		\end{tabular}
	\end{table}

\begin{figure}[ht!]
		\centering
		\includegraphics[width = 0.99\linewidth]{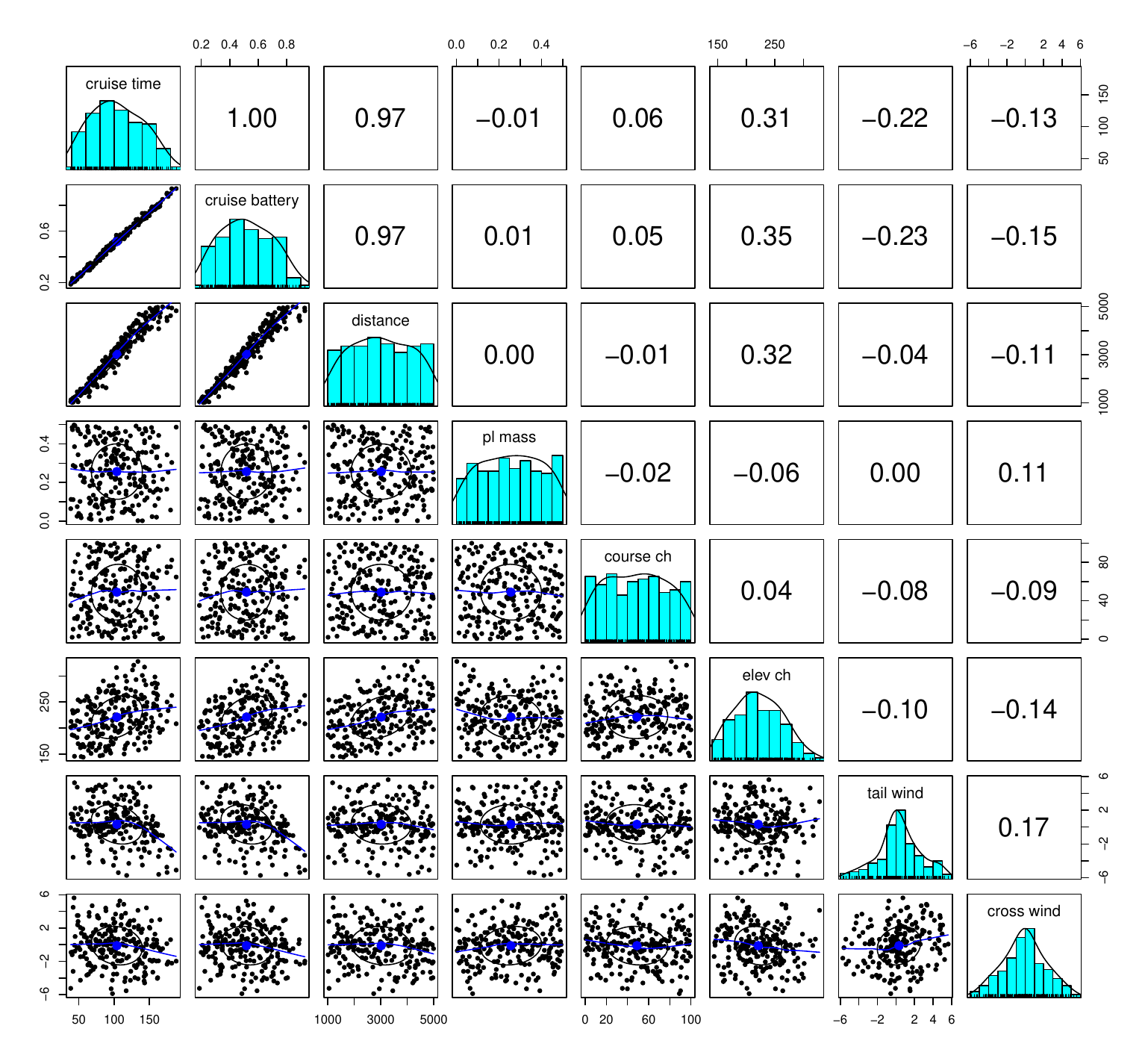}
		\caption{Scatterplot and correlation plot for cruise models.}
		\label{fig:corr:cruise}
	\end{figure}

	We present a scatterplot diagram for the cruise time and cruise battery consumption in \cref{fig:corr:cruise}. We notice that very strong correlation is present between the inputs and the response variables. As a result a significant proportion of variance is explained by the linear model. We present the model selection table for cruise time in \cref{tab:cv_cruise} and cruise battery consumption in \cref{tab:cv_batt_c}.
	
	\begin{table}[ht]
		\centering
		\caption{Ten-fold cross-validation results for cruise time surrogate models.}
		\label{tab:cv_cruise}
		\begin{tabular}{lrrrrrr}
			\toprule
			Model 
			& {$R^2$ Mean} 
			& {$R^2$ SD} 
			& {RMSE Mean} 
			& {RMSE SD} 
			& {MAE Mean} 
			& {MAE SD} \\
			\midrule
			Baseline & -0.046 & 0.072 & 0.379 & 0.039 & 0.313 & 0.033 \\ 
			LM1 & \textbf{0.943} & \textbf{0.025} & \textbf{0.086} & \textbf{0.015} & \textbf{0.064} & \textbf{0.010} \\ 
			GP1 & -0.046 & 0.072 & 0.379 & 0.039 & 0.313 & 0.033 \\ 
			GP2 & -1.850 & 1.251 & 0.604 & 0.102 & 0.502 & 0.100 \\ 
			GP3 & -0.046 & 0.072 & 0.379 & 0.039 & 0.313 & 0.033 \\ 
			GP4 & -1.875 & 1.302 & 0.606 & 0.105 & 0.501 & 0.102 \\ 
			\bottomrule
		\end{tabular}
	\end{table}

	\begin{table}[ht]
		\centering
		\caption{Ten-fold cross-validation results for cruise battery surrogate models.}
		\label{tab:cv_batt_c}
		\begin{tabular}{lrrrrrr}
			\toprule
			Model 
			& {$R^2$ Mean} 
			& {$R^2$ SD} 
			& {RMSE Mean} 
			& {RMSE SD} 
			& {MAE Mean} 
			& {MAE SD} \\
			\midrule
			Baseline & -0.043 & 0.065 & 0.377 & 0.040 & 0.313 & 0.034 \\ 
			LM1 & \textbf{0.941} & \textbf{0.024} & \textbf{0.088} & \textbf{0.016} & \textbf{0.066} & \textbf{0.009} \\ 
			GP1 & -0.043 & 0.065 & 0.377 & 0.040 & 0.313 & 0.034 \\ 
			GP2 & 0.277 & 0.268 & 0.307 & 0.049 & 0.250 & 0.046 \\ 
			GP3 & -0.043 & 0.065 & 0.377 & 0.040 & 0.313 & 0.034 \\ 
			GP4 & 0.273 & 0.274 & 0.307 & 0.049 & 0.251 & 0.046 \\ 
			\bottomrule
		\end{tabular}
	\end{table}
	
	{
    \section{Additional Plots and Results}
    \subsection{Independent vs joint probability}

    Here we present the analysis between independent and joint coverage probability. For joint coverage probability we consider 200 Monte Carlo samples to compute the joint probability for each case. We presented the result in \cref{fig:coverage:compare}. We see that both independent and joint probability are along the diagonal suggesting very nominal differences. Moreover, we also see that independent product is actually conservative which we can see from the figure in bottom panel.

    \begin{figure}
        \centering
        \includegraphics[width=0.95\linewidth]{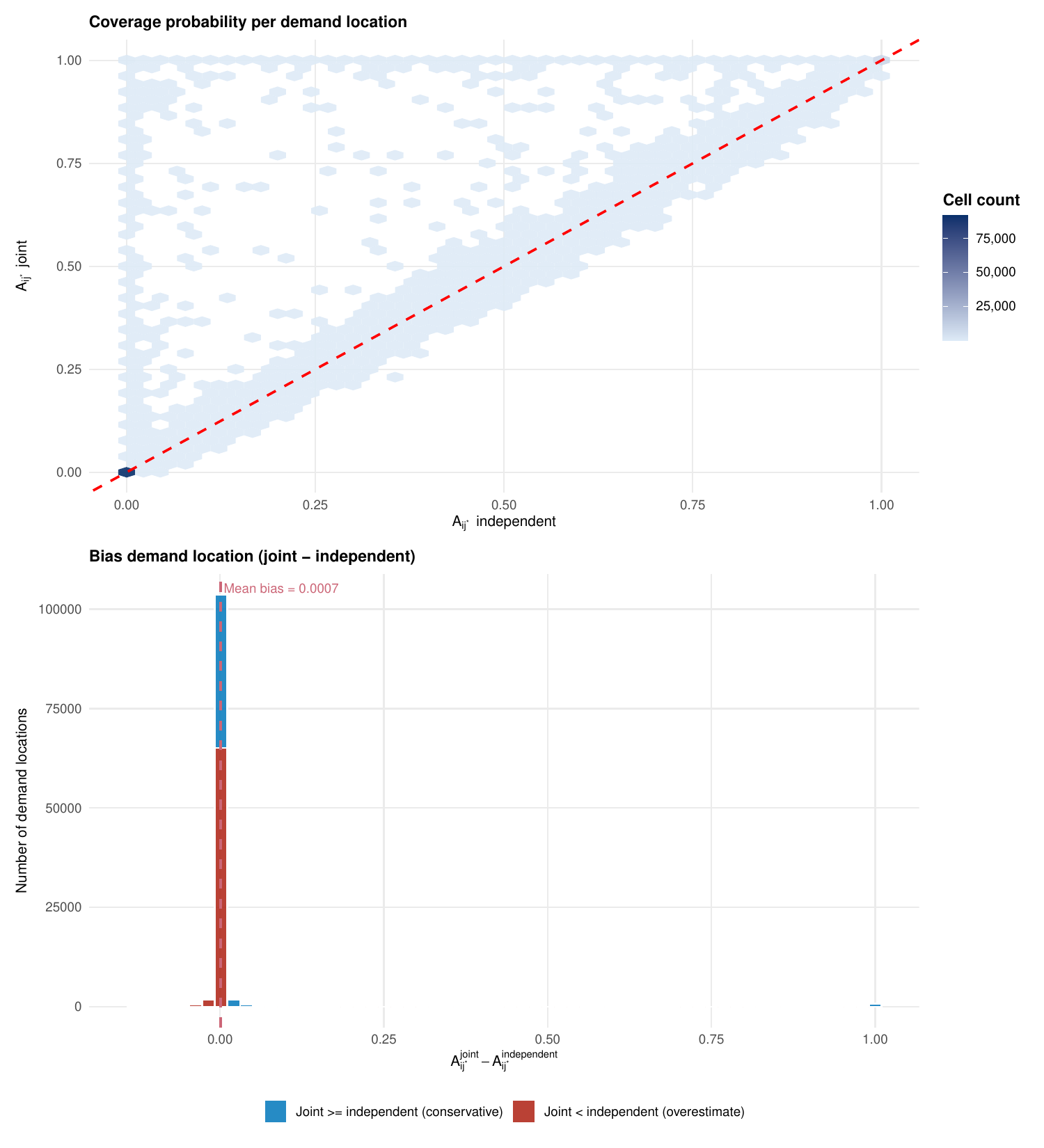}
        \caption{Joint vs Independent coverage probability.}
        \label{fig:coverage:compare}
    \end{figure}

	\subsection{Sensitivity Analysis with Hyperparameters}
	In \cref{tab:sensti} we have presented the sensitivity analysis over different parameter values. The metrics are obtained by fixing the parameter of interest and averaging over all possible combinations.
	\begin{table}[ht]
		\centering
		\caption{Median estimates with respect to different hyper parameters presented in the model. Total 450 combinations are considered for the sensitivity analysis involving GGC area and for temperature parameter $\beta = 15$ is fixed. For the sake of computation 8 equidistant hours of the day along with all seasons are chosen with 500 iterations for MCMC run. Estimates are obtained by discarding first 100 simulations.}
		\label{tab:sensti}
		\begin{tabular}{c|ccccc}
			\toprule
			& \multicolumn{5}{c}{Location parameter of logistic transformation ($a$)}\\
			\midrule
			Test & Value & No.~Sites & Cases & QALY & Cost per QALY \\ 
			\midrule
			1 & 1 & 13 & 550 & 93 & 4612 \\ 
			2 & 2 & 14 & 553 & 93 & 4727 \\ 
			3 & 3 & 15 & 555 & 94 & 4783 \\ 
			4 & 4 & 15 & 556 & 94 & 4843 \\ 
			5 & 5 & 15 & 556 & 94 & 4880 \\ 
			\midrule
			& \multicolumn{5}{c}{Scale parameter of logistic transformation ($b$)}\\
			\midrule
			Test & Value & No.~Sites & Cases & QALY & Cost per QALY \\ 
			\midrule
			1 & 5 & 14 & 550 & 93 & 4626 \\ 
			2 & 10 & 14 & 553 & 94 & 4724 \\ 
			3 & 15 & 15 & 554 & 94 & 4787 \\ 
			4 & 20 & 15 & 556 & 94 & 4835 \\ 
			5 & 25 & 15 & 556 & 94 & 4873 \\ 
			\hline
			\midrule
			& \multicolumn{5}{c}{Weight for the second utility ($\eta$)}\\
			\midrule
			Test & Value & No.~Sites & Cases & QALY & Cost per QALY \\ 
			\midrule
			1 & 0.0 & 14 & 551 & 92 & 4708 \\ 
			2 & 0.2 & 14 & 553 & 93 & 4708 \\ 
			3 & 0.4 & 14 & 554 & 94 & 4727 \\ 
			4 & 0.6 & 15 & 555 & 94 & 4766 \\ 
			5 & 0.8 & 15 & 556 & 94 & 4830 \\ 
			6 & 1.0 & 15 & 556 & 95 & 4875 \\ 
			\midrule
			& \multicolumn{5}{c}{Equity parameter ($\gamma$)}\\
			\midrule
			Test & Value & No.~Sites & Cases & QALY & Cost per QALY \\ 
			\midrule
			1 & 0.0 & 15 & 554 & 94 & 4764 \\ 
			2 & 0.5 & 15 & 554 & 94 & 4767 \\ 
			3 & 1.0 & 15 & 554 & 94 & 4776 \\
			\bottomrule
		\end{tabular}
	\end{table}
}

	\section{Additional Results for Case Studies}
\begin{figure}[ht!]
		\centering
		\includegraphics[width = 0.99\linewidth]{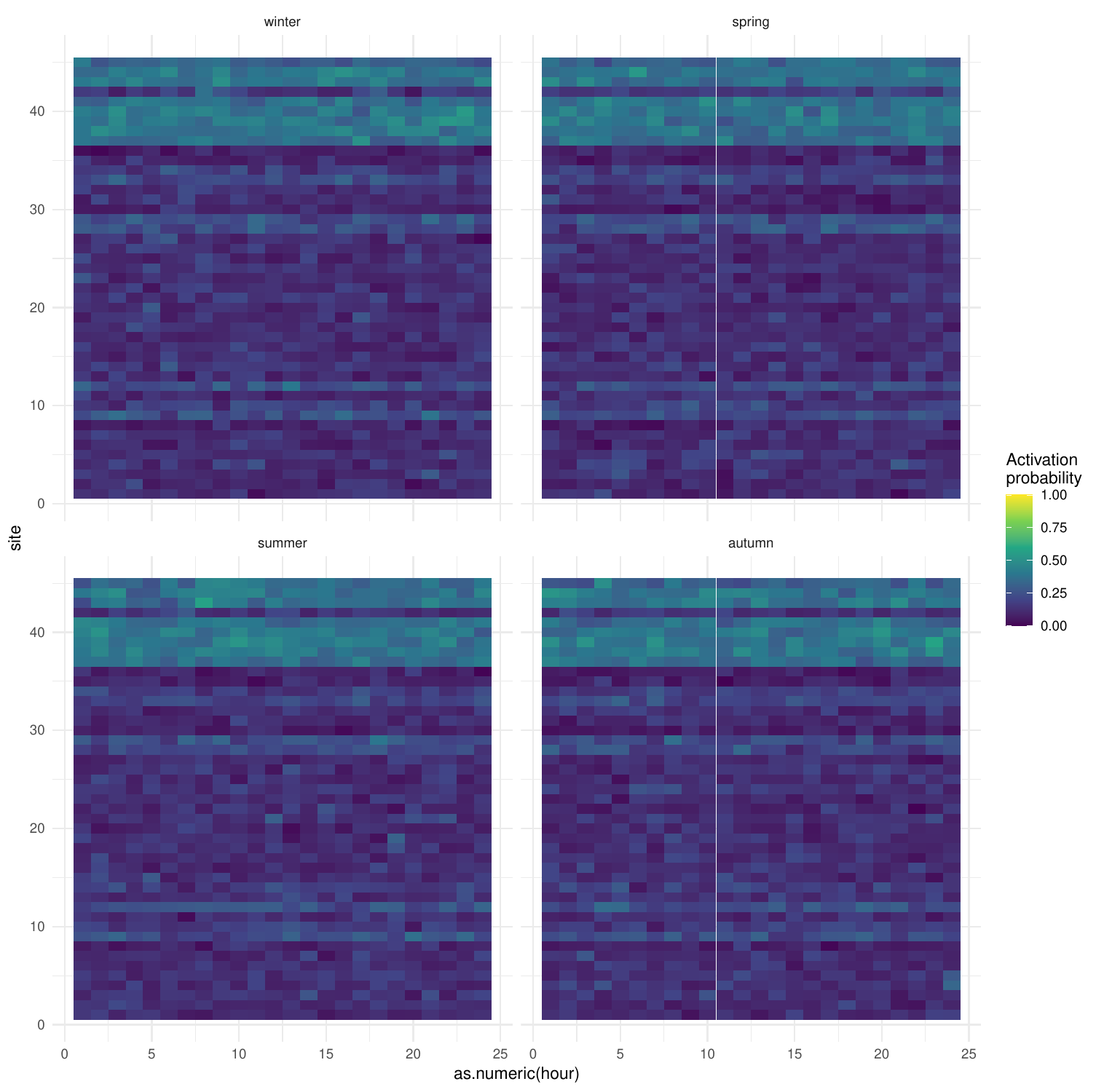}
		\caption{Posterior activation probability across different season hour combination for NHS GGC.}
		\label{fig:active:ggc}
	\end{figure}

\begin{figure}[ht!]
		\centering
		\includegraphics[width = 0.99\linewidth]{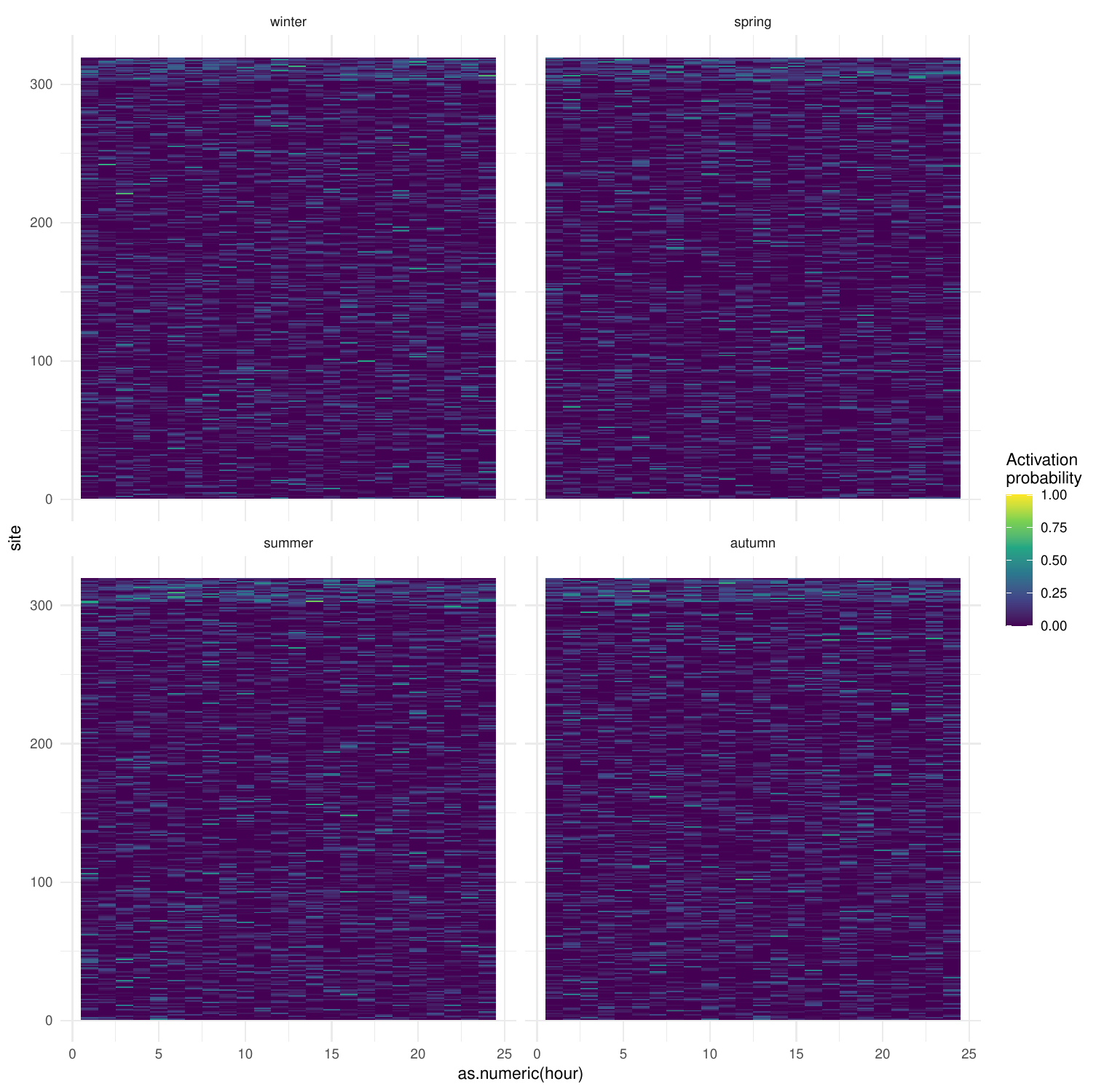}
		\caption{Posterior activation probability across different season hour combination for NHS Grampian.}
		\label{fig:active:grampian}
	\end{figure}

	The activation plot for selected network design of NHS GGC is presented in \cref{fig:active:ggc}. Here the `x'-axis denotes the hour of the day and `y'-axis denote the site id. The top 9 rows are existing SAS stations and as expected, they are likely to be selected irrespective of the seasons. The new locations tend to fluctuate more that can also be notice. Similarly, we present the activation plot for NHS Grampian in \cref{fig:active:grampian} where the top 17 rows are existing SAS stations.

	Diagnostic summary of MCMC convergence is summarised for NHS GGC  in \cref{tab:s1_mcmc_diagnostics} and for NHS Grampian in \cref{tab:s1_mcmc_diagnostics_grampian}. Note that while the effective sample size may seem low but it is worth noting that there are total 319 possible locations and the combinatorial space becomes $2^{319}$. Additionally we have given a traceplot for all season for the best network at 12:00 pm in

	\begin{table}[ht]
		\centering
		\scriptsize
		\caption{MCMC convergence diagnostics for NHS GGC across all temperature
			parameter values $\beta$ and seasons.}
		\label{tab:s1_mcmc_diagnostics}
		\vspace{4pt}
		\resizebox{\textwidth}{!}{%
			\begin{tabular}{cc cc cc cc c r}
				\toprule
				& & \multicolumn{2}{c}{ESS log-posterior} & \multicolumn{2}{c}{ESS network size}
				& \multicolumn{2}{c}{Acceptance rate} & Collapsed & Moves \\
				$\beta$ & Season & Mean & Median & Mean & Median
				& Mean & Median & (frac.) & (mean) \\
				\midrule
				5 & autumn & 71.29 & 70.00 & 77.69 & 77.69 & 0.33 & 0.33 & 0.00 & 1321.46 \\ 
				 & spring & 69.20 & 68.57 & 74.09 & 72.32 & 0.33 & 0.33 & 0.00 & 1319.46 \\ 
				 & summer & 67.67 & 68.13 & 75.10 & 73.61 & 0.33 & 0.33 & 0.00 & 1325.25 \\ 
				 & winter & 68.64 & 65.89 & 74.36 & 71.95 & 0.33 & 0.33 & 0.00 & 1313.88 \\ 
				 \midrule
				7 & autumn & 72.00 & 68.48 & 77.96 & 77.01 & 0.33 & 0.34 & 0.00 & 1334.21 \\ 
				& spring & 68.67 & 66.20 & 73.25 & 72.75 & 0.33 & 0.33 & 0.00 & 1331.12 \\ 
				& summer & 67.79 & 67.28 & 76.41 & 73.30 & 0.33 & 0.34 & 0.00 & 1338.38 \\ 
				& winter & 68.75 & 66.78 & 75.06 & 72.51 & 0.33 & 0.33 & 0.00 & 1327.75 \\ 
				\midrule
				9& autumn & 72.01 & 69.11 & 78.35 & 77.38 & 0.34 & 0.34 & 0.00 & 1348.50 \\ 
				& spring & 68.68 & 67.23 & 74.53 & 73.39 & 0.34 & 0.33 & 0.00 & 1343.25 \\ 
				& summer & 67.80 & 67.91 & 76.38 & 73.19 & 0.34 & 0.34 & 0.00 & 1352.38 \\ 
				& winter & 68.04 & 67.22 & 74.90 & 72.78 & 0.34 & 0.34 & 0.00 & 1341.17 \\ 
				\midrule
				11& autumn & 71.18 & 69.69 & 78.54 & 76.97 & 0.34 & 0.34 & 0.00 & 1362.04 \\ 
				& spring & 68.32 & 66.75 & 74.32 & 74.35 & 0.34 & 0.34 & 0.00 & 1356.38 \\ 
				& summer & 67.98 & 68.29 & 76.24 & 73.28 & 0.34 & 0.34 & 0.00 & 1365.58 \\ 
				& winter & 67.81 & 66.20 & 74.50 & 72.49 & 0.34 & 0.34 & 0.00 & 1353.62 \\ 
				\midrule
				13& autumn & 72.12 & 69.84 & 79.84 & 78.67 & 0.34 & 0.34 & 0.00 & 1375.92 \\ 
				& spring & 67.71 & 67.21 & 74.72 & 72.76 & 0.34 & 0.34 & 0.00 & 1368.88 \\ 
				& summer & 67.79 & 67.59 & 76.20 & 74.07 & 0.34 & 0.34 & 0.00 & 1377.79 \\ 
				& winter & 68.56 & 66.62 & 74.79 & 70.66 & 0.34 & 0.34 & 0.00 & 1366.79 \\ 
				\midrule
				15& autumn & 72.33 & 69.53 & 80.01 & 79.89 & 0.35 & 0.35 & 0.00 & 1389.25 \\ 
				& spring & 68.51 & 67.94 & 75.34 & 75.05 & 0.35 & 0.34 & 0.00 & 1380.00 \\ 
				& summer & 67.95 & 68.19 & 77.33 & 74.42 & 0.35 & 0.35 & 0.00 & 1393.58 \\ 
				& winter & 68.40 & 66.51 & 75.69 & 72.57 & 0.34 & 0.34 & 0.00 & 1377.92 \\ 
				\midrule
				17& autumn & 72.31 & 68.76 & 80.31 & 78.86 & 0.35 & 0.35 & 0.00 & 1403.54 \\ 
				& spring & 69.14 & 67.41 & 76.38 & 75.78 & 0.35 & 0.34 & 0.00 & 1391.62 \\ 
				& summer & 68.04 & 67.79 & 77.65 & 75.76 & 0.35 & 0.35 & 0.00 & 1406.42 \\ 
				& winter & 68.82 & 67.17 & 76.21 & 74.47 & 0.35 & 0.35 & 0.00 & 1394.04 \\ 
				\midrule
				19& autumn & 72.89 & 70.47 & 81.62 & 78.97 & 0.35 & 0.36 & 0.00 & 1416.08 \\ 
				& spring & 68.43 & 66.31 & 76.62 & 75.70 & 0.35 & 0.35 & 0.00 & 1405.21 \\ 
				& summer & 68.11 & 67.75 & 77.70 & 74.42 & 0.36 & 0.36 & 0.00 & 1420.79 \\ 
				& winter & 68.58 & 68.88 & 76.49 & 73.35 & 0.35 & 0.35 & 0.00 & 1407.21 \\ 
				\midrule
				21& autumn & 72.46 & 70.68 & 80.87 & 79.14 & 0.36 & 0.36 & 0.00 & 1430.29 \\ 
				& spring & 68.75 & 68.04 & 76.66 & 75.40 & 0.36 & 0.35 & 0.00 & 1419.75 \\ 
				& summer & 68.82 & 68.19 & 79.74 & 78.03 & 0.36 & 0.36 & 0.00 & 1434.25 \\ 
				& winter & 68.02 & 65.18 & 76.86 & 76.13 & 0.36 & 0.36 & 0.00 & 1421.62 \\ 
				\midrule
				23& autumn & 72.71 & 70.07 & 81.14 & 77.97 & 0.36 & 0.36 & 0.00 & 1443.38 \\ 
				& spring & 69.19 & 67.30 & 77.75 & 77.29 & 0.36 & 0.36 & 0.00 & 1432.54 \\ 
				& summer & 68.52 & 66.45 & 79.89 & 78.40 & 0.36 & 0.36 & 0.00 & 1447.46 \\ 
				& winter & 68.93 & 68.01 & 77.37 & 76.21 & 0.36 & 0.36 & 0.00 & 1435.00 \\ 
				\midrule
				25& autumn & 72.76 & 70.16 & 82.11 & 81.80 & 0.36 & 0.37 & 0.00 & 1457.96 \\ 
				& spring & 68.80 & 67.83 & 78.12 & 78.66 & 0.36 & 0.36 & 0.00 & 1445.79 \\ 
				& summer & 69.17 & 67.91 & 79.05 & 77.77 & 0.36 & 0.37 & 0.00 & 1459.17 \\ 
				& winter & 68.24 & 67.16 & 76.74 & 75.05 & 0.36 & 0.36 & 0.00 & 1448.96 \\
				\bottomrule
			\end{tabular}%
		}
	\end{table}
	
	\begin{table}[ht]
		\centering
		\scriptsize
		\caption{MCMC convergence diagnostics for NHS Grampian across all temperature
			parameter values $\beta$ and seasons.}
		\label{tab:s1_mcmc_diagnostics_grampian}
		\vspace{4pt}
		\resizebox{\textwidth}{!}{%
			\begin{tabular}{cc cc cc cc c r}
				\toprule
				& & \multicolumn{2}{c}{ESS log-posterior} & \multicolumn{2}{c}{ESS network size}
				& \multicolumn{2}{c}{Acceptance rate} & Collapsed & Moves \\
				$\beta$ & Season & Mean & Median & Mean & Median
				& Mean & Median & (frac.) & (mean) \\
				\midrule
				5 & autumn & 10.47 &  9.80 & 10.91 &  9.13 & 0.10 & 0.10 & 0.00 & 409.08 \\
				& spring & 10.16 &  9.45 & 10.41 &  9.49 & 0.10 & 0.10 & 0.00 & 410.04 \\
				& summer & 10.76 & 10.76 & 11.27 & 10.46 & 0.10 & 0.10 & 0.00 & 415.13 \\
				& winter & 11.25 & 10.86 & 11.60 & 11.00 & 0.10 & 0.10 & 0.00 & 406.92 \\
				\midrule
				7 & autumn &  9.35 &  8.94 &  9.39 &  8.99 & 0.10 & 0.10 & 0.00 & 413.13 \\
				& spring &  9.09 &  9.07 &  9.06 &  9.10 & 0.10 & 0.10 & 0.00 & 399.71 \\
				& summer & 10.19 &  9.08 &  9.34 &  9.45 & 0.10 & 0.10 & 0.00 & 411.54 \\
				& winter &  9.63 &  9.37 &  9.65 &  9.22 & 0.10 & 0.10 & 0.00 & 415.17 \\
				\midrule
				9 & autumn & 10.38 &  9.97 & 10.54 &  9.64 & 0.10 & 0.10 & 0.00 & 410.83 \\
				& spring & 10.20 &  9.39 &  9.63 &  9.26 & 0.10 & 0.10 & 0.00 & 412.17 \\
				& summer & 10.30 & 10.22 & 11.47 & 10.70 & 0.10 & 0.10 & 0.00 & 417.29 \\
				& winter & 11.22 &  9.68 & 11.51 & 11.11 & 0.10 & 0.10 & 0.00 & 408.63 \\
				\midrule
				11 & autumn &  9.23 &  8.48 &  8.95 &  8.65 & 0.10 & 0.10 & 0.00 & 415.04 \\
				& spring & 10.86 & 10.84 & 10.79 & 10.80 & 0.10 & 0.10 & 0.00 & 401.71 \\
				& summer & 10.18 &  9.43 &  9.11 &  9.42 & 0.10 & 0.10 & 0.00 & 414.58 \\
				& winter &  9.56 &  9.82 &  9.82 &  8.22 & 0.10 & 0.10 & 0.00 & 417.71 \\
				\midrule
				13 & autumn & 10.27 &  9.58 & 10.52 &  9.69 & 0.10 & 0.10 & 0.00 & 412.54 \\
				& spring & 10.29 &  9.58 & 10.52 &  9.61 & 0.10 & 0.10 & 0.00 & 414.29 \\
				& summer & 10.88 & 10.45 & 11.48 &  9.95 & 0.10 & 0.10 & 0.00 & 419.13 \\
				& winter & 11.29 & 10.78 & 11.41 & 11.04 & 0.10 & 0.10 & 0.00 & 410.79 \\
				\midrule
				15 & autumn &  9.21 &  8.89 &  9.23 &  8.91 & 0.10 & 0.10 & 0.00 & 416.58 \\
				& spring &  9.88 &  9.70 & 10.75 &  9.93 & 0.10 & 0.10 & 0.00 & 403.83 \\
				& summer & 10.16 &  9.92 &  9.87 &  8.94 & 0.10 & 0.10 & 0.00 & 415.96 \\
				& winter &  9.51 &  9.17 & 15.32 & 14.84 & 0.10 & 0.10 & 0.00 & 419.17 \\
				\midrule
				17 & autumn & 10.41 &  9.86 & 10.53 & 10.71 & 0.10 & 0.10 & 0.00 & 415.08 \\
				& spring & 10.23 & 10.51 & 10.51 & 10.77 & 0.10 & 0.10 & 0.00 & 416.63 \\
				& summer & 10.90 & 10.92 & 11.56 & 10.91 & 0.10 & 0.10 & 0.00 & 421.50 \\
				& winter & 11.35 & 11.74 & 11.58 & 11.08 & 0.10 & 0.10 & 0.00 & 412.58 \\
				\midrule
				19 & autumn &  9.40 &  9.38 &  9.70 &  9.91 & 0.10 & 0.10 & 0.00 & 416.42 \\
				& spring & 10.19 &  9.67 &  9.68 &  9.83 & 0.10 & 0.10 & 0.00 & 418.92 \\
				& summer &  9.39 &  9.66 &  9.47 &  9.55 & 0.10 & 0.10 & 0.00 & 418.88 \\
				& winter & 11.39 & 10.79 & 11.62 & 12.01 & 0.10 & 0.10 & 0.00 & 413.71 \\
				\midrule
				21 & autumn &  9.39 &  9.80 &  9.56 &  9.52 & 0.10 & 0.10 & 0.00 & 418.96 \\
				& spring & 10.88 & 10.80 & 10.81 & 10.59 & 0.10 & 0.10 & 0.00 & 405.83 \\
				& summer & 10.22 &  9.24 &  9.11 & 10.39 & 0.10 & 0.10 & 0.00 & 418.88 \\
				& winter &  9.56 &  9.56 & 14.90 & 14.90 & 0.11 & 0.11 & 0.00 & 423.38 \\
				\midrule
				23 & autumn &  9.60 &  9.97 & 10.55 & 10.38 & 0.11 & 0.11 & 0.00 & 418.04 \\
				& spring & 13.45 & 13.46 & 12.23 & 11.72 & 0.10 & 0.09 & 0.00 & 414.29 \\
				& summer & 10.82 & 10.24 & 11.49 & 11.39 & 0.11 & 0.11 & 0.00 & 425.04 \\
				& winter & 11.26 & 10.44 & 11.56 & 10.63 & 0.11 & 0.11 & 0.00 & 416.21 \\
				\midrule
				25 & autumn &  9.46 &  9.15 &  9.29 &  8.72 & 0.10 & 0.10 & 0.00 & 420.79 \\
				& spring &  9.74 &  9.18 & 10.26 & 10.26 & 0.10 & 0.09 & 0.00 & 408.33 \\
				& summer & 10.27 &  9.02 & 10.15 &  9.49 & 0.11 & 0.10 & 0.00 & 420.96 \\
				& winter &  9.64 &  9.64 &  9.81 &  9.81 & 0.11 & 0.11 & 0.00 & 425.50 \\
				\bottomrule
			\end{tabular}%
		}
	\end{table}
	
	\begin{figure}[ht!]
		\centering
		\includegraphics[width = 0.99\linewidth]{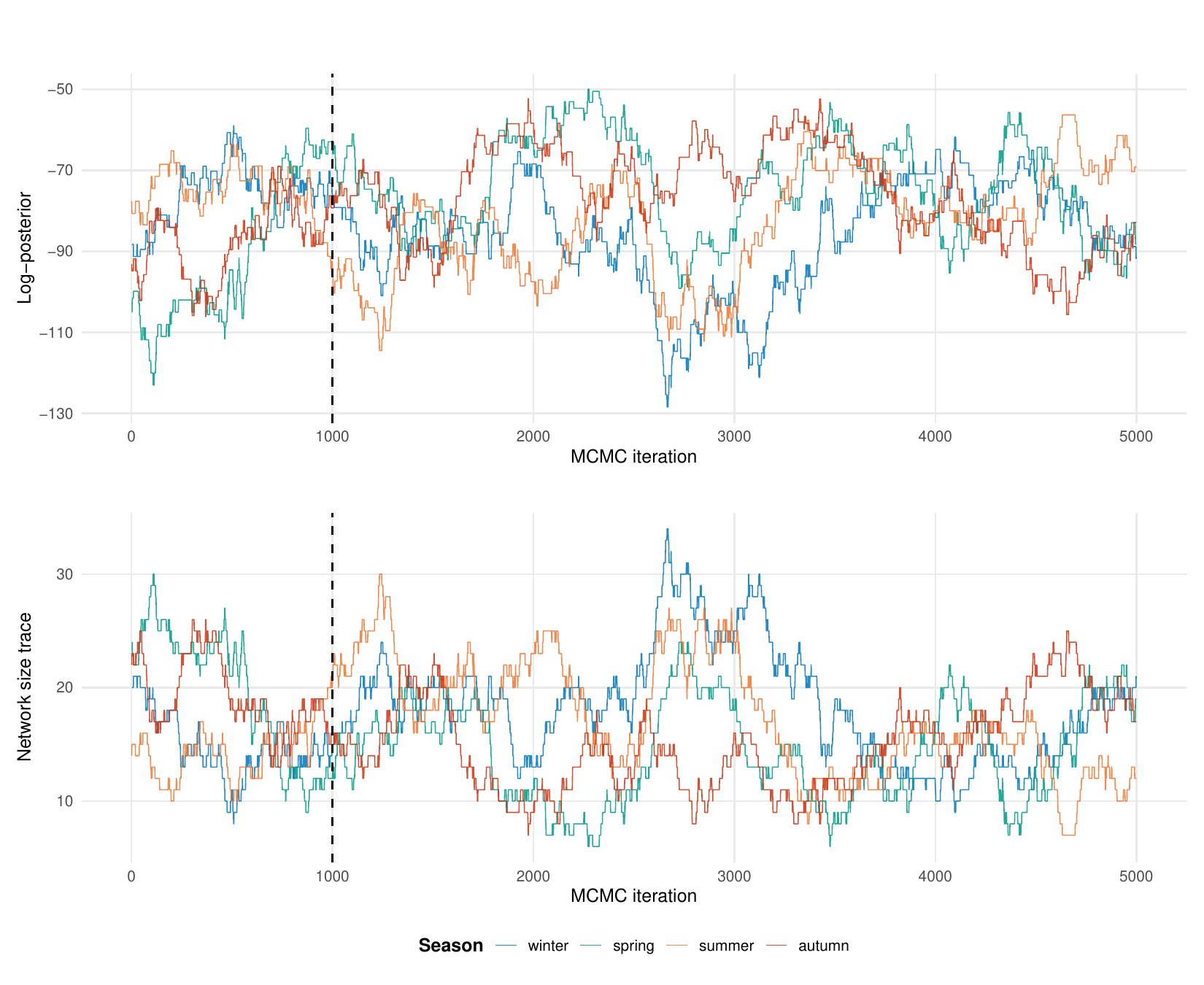}
		\caption{Traceplot of network size and log-posterior for all seasons at the afternoon for NHS Grampian.}
		\label{fig:trace:grampian}
	\end{figure}

	{\section{Tail-Risk Assessment under Extreme Wind}
		\label{app:tail}
		
		\subsection{Approach}
		\label{app:tail:approach}
		
		Since we are interested in the design phase of the network, the network remains fixed and is not reconfigured in case of an adverse event. Instead, we monitor the  robustness against rare and extreme conditions at the design stage. For that, we compute the expected QALY along with a tail-weighted objective under extreme events. We obtain this quantities using the posterior samples and seasonal wind samples.
		
		Let $x\in\{0,1\}^{p}$ denote a design (a set of active sites), let $_s$ denote the index for wind scenario such that $1\le s\le S$, and let $Q_s(x)$ be the QALY gained by the design $x$ under scenario $s$, with expected performance 
        \begin{equation}
            \bar{Q}(x)=S^{-1}\sum_s Q_s(x).
        \end{equation}
        Lastly, let $x^{\star}$ be the cost-effective design selected by tuning over a set of $\beta$.
		
		\paragraph{Value at risk}
		Since QALY is perceived as a reward, we consider the \emph{lower} tail of $Q_s(x)$. For a predefined tail level $\alpha$ $(=0.05)$, we define the value-at-risk as the lower $\alpha$-quantile such that
        \begin{equation}
            \mathrm{VaR}_{\alpha}(x)=\inf\left\{t:\Pr\left[Q_s(x)\le t\right]\ge\alpha\right\}.
        \end{equation}
		and the conditional value-at-risk as the mean QALY over the worst $\alpha$ fraction of scenarios,
		\begin{equation}
			\mathrm{CVaR}_{\alpha}(x)=\frac{1}{\alpha}
			\mathbb{E}\left[Q_s(x)\mathbb{I}\left(Q_s(x)\le\mathrm{VaR}_{\alpha}(x)\right)\right].
			\label{eq:cvar}
		\end{equation}
		This way can capture the sensitivity to the shape of the lower tail and unlike a single worst-case statistic it averages over the tail which allows us to monitor robustness of the design.
		
		\paragraph{Value at risk under design uncertainty}
		Note that using \cref{eq:cvar} we can easily compute the conditional-value-at-risk of the final design $x^\star$ to showcase the aleatory uncertainty due to wind. However, a distinctive feature of the Gibbs-posterior formulation is that we can also quantify the epistemic uncertainty over the design itself. The Gibbs posterior
        \begin{equation}
            \Pi(x)\propto\exp\{-\beta\,\ell(x)\}\pi(x)
        \end{equation}
        places mass on a set of admissible configurations rather than a single point which allows us to evaluate two additional metrics of robustness. First, we define the \emph{posterior-predictive} CVaR as the lower-tail CVaR of the mixture $F_{\mathrm{pred}}(t)=\mathbb{E}_{x\sim\Pi}[\Pr(Q_s(x)\le t)]$. Moreover, we consider the \emph{posterior distribution of the tail statistic}, where we evaluate $\mathrm{CVaR}_{\alpha}(x)$ as a functional of $x$ and compute the posterior mean and $95\%$ credible interval. This way, we can compute an explicit measure of confidence in the reported tail number that point-estimate baselines cannot provide.

		\subsection{Results}
		\label{app:tail:results}

        Here, for the sake of brevity, we only present our analyses with NHS GGC. As shown in the main manuscript, for NHS GGC, we have only 8 sites. We notice that the selected network performs well even in the extreme conditions. Its worst $5\%$ coverage is $\mathrm{CVaR}_{0.05}=62.59$ QALY gained, retaining approximately $80\%$ of the expected performance. We also notice that the value-at-risk and conditional value-at-risk lie close together ($65.55$ versus $62.59$), indicating a thin lower tail without a catastrophic outcome. We present these metrics in the upper panel of \cref{tab:cvar}.
		
		\begin{table}[ht]
			\centering
			\caption{Tail-weighted performance of the network design for NHS GGC and $\alpha=0.05$). The upper panel reports tail statistics under aleatory / wind uncertainty for the final cost-effective final design and the lower panel propagates the Gibbs posterior over designs through the tail functional.}
			\label{tab:cvar}
			\begin{tabular}{lcccc}
				\toprule
				Quantity & $\bar{Q}$ & $\mathrm{VaR}_{0.05}$ & $\mathrm{CVaR}_{0.05}$
				& $\mathrm{CVaR}/\bar{Q}$\\
				\midrule
				\multicolumn{5}{l}{\emph{Final design (aleatory uncertainty)}}\\
				Cost-effective design $x^{\star}$ ($8$ sites)
				& $78.58$ & $65.55$ & $62.59$ & $0.80$\\
				\midrule
				\multicolumn{5}{l}{\emph{Under design uncertainty (posterior of $x$)}}\\
				Posterior-predictive $\mathrm{CVaR}_{0.05}$
				& --- & --- & $22.44$ & ---\\
				Posterior mean $\mathrm{CVaR}_{0.05}$ ($95\%$ CrI)
				& --- & --- & $47.78$ $[12.87,\,65.58]$ & ---\\
				\bottomrule
			\end{tabular}
		\end{table}
		
		\begin{figure}[htbp]
			\centering
			\includegraphics[width=0.72\textwidth]{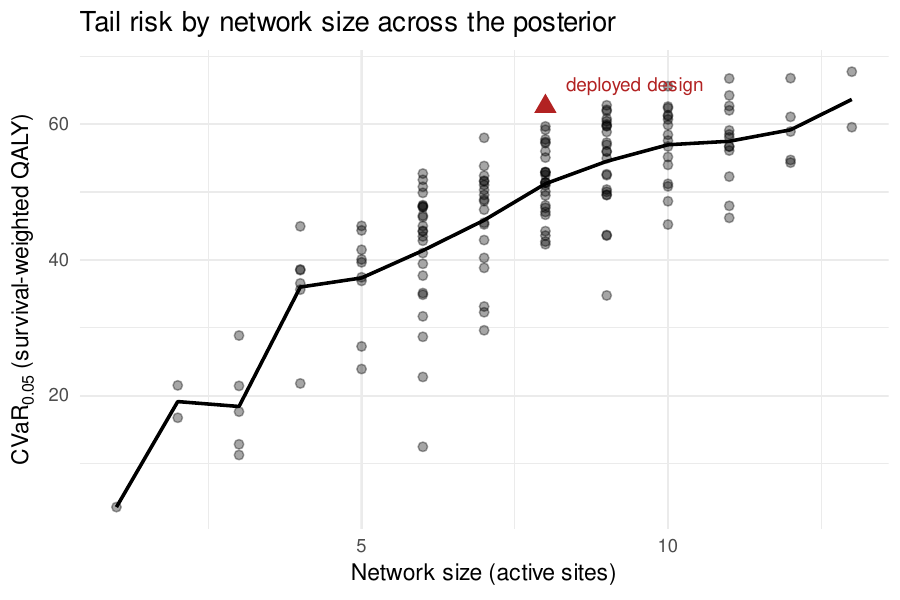}
			\caption{Tail risk against network size across the design posterior for NHS GGC  and $\alpha=0.05$. Each grey point is one posterior design draw and the black line represents the posterior-weighted mean $\mathrm{CVaR}_{0.05}$ for each network size. The final cost-effective design ($8$ sites) is marked with red triangle.}
			\label{fig:cvar_size}
		\end{figure}
		
		We notice a very different behaviour when we consider the design uncertainty which we present in \cref{tab:cvar}. The posterior-predictive $\mathrm{CVaR}_{0.05}=22.44$ QALY and the posterior mean $\mathrm{CVaR}_{0.05}=47.78$ QALY with a wide $95\%$ credible interval $[12.87,\,65.58]$ all fall well below the $\mathrm{CVaR}$ obtained for the final cost-effective design. This is a direct consequence of the Gibbs posterior placing mass across network configurations of very different scales that range from $1$ to $13$ active sites. We show the behaviour of the posterior $\mathrm{CVaR}$ in \cref{fig:cvar_size}. We notice that the tail QALY gain increases steeply up to roughly seven sites and then slows down. Therefore, the tail robustness exhibits a diminishing returns in network size suggesting the importance of designing a network which is not over saturated and does not incur cost without actually improving the coverage. From \cref{fig:cvar_size}, we notice that design attains $\mathrm{CVaR}_{0.05}$ near the upper edge of its own size band of 8 suggesting that our proposed design is the best possible solution if we consider only 8 drone stations. Moreover, the $\mathrm{CVaR}$ is close to that of substantially larger networks validating the cost-effectiveness trade-off that we consider for selecting the best network.
		
		Note that the gap between the two panels of \cref{tab:cvar} reflects an important benefit of a Bayesian learning paradigm. The final design obtained through our learning is equivalent to deterministic facility-location or minimax solution which returns a single network and a single tail number, conveying no information about how sensitive that number is to the design uncertainty itself. The Gibbs-posterior framework can answer these two questions. We can calculate the tail risk of the selected network that can be deployed ($\mathrm{CVaR}_{0.05}=62.59$), as well as the tail risk under design uncertainty. The difference between these two metrics quantifies how much of the resilience of the network is affected due to design choice, an epistemic diagnosis that is unavailable to standard point estimate methods. However, it is imperative to emphasise that the predictive statistic reflects uncertainty over network \emph{size} across the posterior support and not the operational risk of the final design.}
        
\bibliographystyle{elsarticle-harv} 
\bibliography{example}






\end{document}